\documentclass[a4paper]{report}
\usepackage[utf8]{inputenc}
\usepackage[T1]{fontenc}
\usepackage{RJournal}
\usepackage{amsthm,amsmath,amssymb,amstext} 
\usepackage{booktabs}
\usepackage{array}

\usepackage{nicefrac}
\usepackage{caption}
\usepackage{tabularx}
\usepackage{graphicx}
\usepackage{booktabs}
\usepackage{lscape}
\usepackage{placeins}
\usepackage{colortbl}
\usepackage{dsfont}
\usepackage{multirow}
\usepackage{listings}
\usepackage{adjustbox}
\usepackage{setspace}

\begin{document}

\sectionhead{Contributed research article}
\volume{XX}
\volnumber{YY}
\year{2017}
\month{AAAA}

\begin{article}
  
\title{Multilabel Classification with R Package \pkg{mlr}}
\author{by Philipp Probst, Quay Au, Giuseppe Casalicchio, Clemens Stachl and Bernd Bischl}

\maketitle

\abstract{
We implemented several multilabel classification algorithms in the machine 
learning package \CRANpkg{mlr}. The implemented methods are binary relevance, classifier chains,
nested stacking, dependent binary relevance and stacking, which can be used with any base learner that is accessible in \pkg{mlr}. 
Moreover, there is access to the multilabel classification versions of \CRANpkg{randomForestSRC} and \CRANpkg{rFerns}. 
All these methods can be easily compared by different implemented multilabel performance measures and resampling methods in the standardized \pkg{mlr} framework. 
In a benchmark experiment with several multilabel datasets, the performance of the different methods is evaluated. 
}

\section{Introduction}

Multilabel classification is a classification problem where multiple target labels can be assigned to each observation instead of only one, 
like in multiclass classification. 
It can be regarded as a special case of multivariate classification or multi-target prediction problems, 
for which the scale of each response variable can be of any kind, for example nominal, ordinal or interval. 

Originally, multilabel classification was used for text classification \citep{McCallum1999, Schapire2000} and is now used in several applications in different 
research fields. For example, in image classification, 
a photo can belong to the classes \textit{mountain} and \textit{sunset} simultaneously. 
\citet{Zhang2008} and others \citep{Boutell20041757} used multilabel algorithms to classify scenes on images of natural environments. 
Furthermore, gene functional classifications is a popular application of multilabel learning in the field of biostatistics \citep{NIPS2001_1964, Zhang2008}. 
Additionally, multilabel classification is useful to categorize audio files. Music genres \citep{Sanden2011}, instruments \citep{Kursa2014}, 
bird sounds \citep{briggs2013new} or even emotions evoked by a 
song \citep{trohidis2008multi} can be labeled with several categories. 
A song could, for example, be classified both as a \textit{rock song} and a \textit{ballad}.  

An overview of multilabel classification was given by \citet{Tsoumakas2007}. 
Two different approaches exist for multilabel classification. On the one hand, there are algorithm adaptation methods that try to adapt multiclass algorithms 
so they can be applied directly to the problem.
On the other hand, there are problem transformation methods, which try to transform the multilabel classification into binary or multiclass classification problems. 

Regarding multilabel classification software, there is the \CRANpkg{mldr} \citep{Charte2015} R package that contains some functions to get basic characteristics of specific multilabel datasets. 
The package is also useful for transforming multilabel datasets that are typically saved as ARFF-files (Attribute-Relation File Format) to data frames and vice versa. This is especially helpful 
because until now only the software packages MEKA \citep{Read2012} and Mulan \citep{Tsoumakas2011} were available for multilabel classification and both require multilabel 
datasets saved as ARFF-files to be executed. 
Additionally, the \CRANpkg{mldr} package provides a function that applies the binary relevance or label powerset transformation method which transforms a multilabel dataset into several binary datasets (one for each label) or into a  multiclass dataset using the set of labels for each observation as a single target label, respectively.
However, there is no R package that provides a standardized interface for executing different multilabel 
classification algorithms. With the extension of the \pkg{mlr} package described in this paper, it will be possible to execute several multilabel classification algorithms in R with many different base learners. 

In the following section of this paper, we will describe the implemented multilabel classification methods and then give a practical instruction of how to execute these algorithms in \pkg{mlr}. 
Finally, we present a benchmark experiment that compares the performance of all implemented methods on several datasets. 

\section{Multilabel classification methods implemented in \pkg{mlr}}

In this section, we present multilabel classification algorithms that are implemented in the \pkg{mlr} package \citep{Bischl2016}, which is a powerful and 
mo\-dularized toolbox for machine learning 
in R. The package offers a unified interface to more than a hundred learners from the areas classification, regression, cluster analysis 
and survival analysis. Furthermore, the package provides functions and tools that facilitate complex workflows such as hyperparameter tuning
\citep[see, e.g.,][]{lang2015} and feature selection that can now also be applied to the multilabel classification methods presented in this paper.
In the following, we list the algorithm adaptation methods and problem transformation methods that are currently available in \pkg{mlr}. 

\subsection{Algorithm adaptation methods}

The \pkg{rFerns} \citep{Kursa2014} package contains an extension of the random ferns algorithm for multilabel classification. 
In the \pkg{randomForestSRC} \citep{Ishwaran2013} package, multivariate classification and regression random forests can be created. 
In the classification case, the difference to standard random forests is that a composite normalized Gini index splitting rule is used. 
Multilabel classification can be achieved by using binary encoding for the labels.  

\subsection{Problem transformation methods}

Problem transformation methods try to transform the multilabel classification problem so that a simple binary classification algorithm, the 
so-called base learner, can be applied. 

Let $n$ be the number of observations, let $p$ be the number of predictor variables and let $Z = \left\lbrace z_1, \hdots , z_m \right\rbrace$ be the set of all labels. Observations follow an unknown probability distribution $\mathcal{P}$ on $\mathcal X \times \mathcal Y$, where $\mathcal X$ is a $p-$dimensional input space of arbitrary measurement scales and $\mathcal Y = \lbrace 0,1 \rbrace^m$ is the target space.  
In our notation, $\mathbf x^{(i)} = \left(x_1^{(i)}, \hdots, x_p^{(i)}\right)^\top \in \mathcal X$ refers to the $i$-th observation and $ \mathbf x_{j} = \left(x_j^{(1)}, \hdots, x_j^{(n)}\right)^\top$ refers to the $j$-th predictor variable, for all $i = 1,\hdots,n$ and $j = 1,\hdots,p$. 
The observations $\mathbf x^{(i)}$ are associated with their multilabel outcomes $\mathbf y^{(i)} = \left(y_1^{(i)},\hdots,y_m^{(i)}\right)^\top \in \mathcal Y$, for all $i = 1,\hdots,n$. 
For all $k = 1, \hdots, m$, setting $y_k^{(i)}=1$ indicates the relevance, i.e., the occurrence, of label $z_k$ for observation $\mathbf x^{(i)}$ and setting $y_k^{(i)}=0$ indicates the irrelevance of label $z_k$ for observation $\mathbf x^{(i)}$. The set of all instances thus 
becomes $D = \left\lbrace\left(\mathbf x^{(1)},\mathbf y^{(1)}\right) , \left(\mathbf x^{(2)},\mathbf y^{(2)}\right),\hdots,\left(\mathbf x^{(n)},\mathbf y^{(n)}\right) \right\rbrace$.
Furthermore, $\mathbf y_k = \left(y_k^{(1)},\hdots,y_k^{(n)}\right)^\top $ refers to the $k$-th target vector, for all $k = 1,\hdots,m$. Throughout this paper, we visualize multilabel classification problems in the form of tables ($n=6$, $p=3$, $m=3$):

\begin{equation}
D \hspace{2pt} \hat= \hspace{2pt} 
\scalebox{.55}{
\begin{tabular}{|lll|@{}|@{}|@{}|@{}|@{}|@{}|@{}|@{}|lll|}
\hline
$\mathbf x_1$ & $\mathbf x_2$ & $\mathbf x_3$ & $\mathbf y_1$ & $\mathbf y_2$ & $\mathbf y_3$ \\
\hline
\cellcolor{black!30} & \cellcolor{black!30} & \cellcolor{black!30} & \cellcolor{red!70} 0 & \cellcolor{red!70} 0 & \cellcolor{red!70} 1 \\
\cellcolor{black!30} & \cellcolor{black!30} & \cellcolor{black!30} & \cellcolor{red!70} 1 & \cellcolor{red!70} 0 & \cellcolor{red!70} 1 \\
\cellcolor{black!30} & \cellcolor{black!30} & \cellcolor{black!30} & \cellcolor{red!70} 1 & \cellcolor{red!70} 1 & \cellcolor{red!70} 0 \\
\cellcolor{black!30} & \cellcolor{black!30} & \cellcolor{black!30} & \cellcolor{red!70} 1 & \cellcolor{red!70} 1 & \cellcolor{red!70} 1 \\
\cellcolor{black!30} & \cellcolor{black!30} & \cellcolor{black!30} & \cellcolor{red!70} 1 & \cellcolor{red!70} 1 & \cellcolor{red!70} 0 \\
\cellcolor{black!30} & \cellcolor{black!30} & \cellcolor{black!30} & \cellcolor{red!70} 1 & \cellcolor{red!70} 1 & \cellcolor{red!70} 0 \\
\hline
\end{tabular}
}
\label{eq:truevspred}
\end{equation}

The entries of $\mathbf x_1, \mathbf x_2, \mathbf x_3$ can be of any (valid) kind, like continuous, binary, or categorical. The table in (\ref{eq:truevspred}) visualizes 
this as an empty gray background. The target variables are indicated by a red background and can only take the binary values $0$ or $1$. 

\subsubsection{Binary relevance}

The binary relevance method (BR) is the simplest problem transformation method. BR learns a binary classifier 
for each label. Each classifier $C_1 , \hdots, C_m$ is responsible for predicting the \textit{relevance} of their corresponding label by a $0/1$ prediction:
$$
C_k : \mathcal X \longrightarrow \lbrace 0,1 \rbrace, \quad k = 1, \hdots, m
$$
These binary prediction are then combined to a multilabel target. An unlabeled observation $\mathbf x^{(l)}$ is assigned the prediction $\left(C_1\left(\mathbf x^{(l)}\right),C_2\left(\mathbf x^{(l)}\right),\hdots,C_m\left(\mathbf x^{(l)}\right)\right)^\top$. 
Hence, labels are predicted independently of each other and label dependencies are not taken into account. 
BR has linear computational complexity with respect to the number of labels and can easily be parallelized.

\subsubsection{Modeling label dependence}

In the problem transformation setting, the arguably simplest way \citep{Montanes2014} to model label dependence is to condition classifier models not only on 
$\mathcal X$, but also on other label information. The idea is to augment the input space $\mathcal X$ with information of the output space $\mathcal Y$, 
which is available in the training step. There are different ways to realize this idea of augmenting the input space. In essence, they can be distinguished in the following way:

\begin{itemize}
\item Should the true label information be used? (True vs. predicted label information)
\item For predicting one label $z_k$, should all other labels augment the input space, or only a subset of labels? (Full vs. partial conditioning)
\end{itemize} 

\subsubsection{True vs. predicted label information}
\label{sec:True-vs-Pred}
During the training of a classifier $C_k$ for the label $z_k$, the label information of other labels are available in the training data.
Consequently, these true labels can directly be used as predictors to train the classifier. Alternatively, the predictions that are produced by some classifier can be used instead of the 
true labels. 

A classifier, which is trained on additional labels as predictors, needs those additional 
labels as input variables. Since these labels are not available at prediction time, they need to be predicted first. When the true label information is used to augment 
the feature space in the training of a classifier, the assumption that the training data and the test data should be identically distributed is violated \citep{Senge2013}. 
If the true label information is used in the training data and the predicted label information is used in the test data, the training data is not representative for the 
test data. However, experiments \citep{Montanes2014, Senge2013} show that none of these methods should be dismissed immediately. 
Note that we use the superscript ``true'' or ``pred'' to emphasize that a classifier $C_k^\text{true}$ or $C_k^\text{pred}$ used true labels or predicted labels as additional predictors during 
training, respectively. 

Suppose there are $n=6$ observations with $p = 3$ predictors and $m = 3$ labels. The true label $\mathbf y_3$ shall be used to augment the feature space of a binary 
classifier $C_1^\text{true}$ for label $\mathbf y_1$. $C_1^\text{true}$ is thus trained on all predictors and the true label $\mathbf y_3$. The binary classification task for label $\mathbf y_1$ is therefore:
\begin{equation}
\text{Train } C_1^\text{true} \text{ on } \hspace{2pt}
\scalebox{.55}{
\begin{tabular}{|llll|@{}|@{}|@{}|@{}|@{}|@{}|@{}|@{}|l|}
\hline
$\mathbf x_1$ & $\mathbf x_2$ & $\mathbf x_3$  & $\mathbf y_3$ & $\mathbf y_1$ \\ 
\hline
\cellcolor{black!30} & \cellcolor{black!30} & \cellcolor{black!30} & \cellcolor{black!50} 0 & \cellcolor{red!70} 0 \\ 
\cellcolor{black!30} & \cellcolor{black!30} & \cellcolor{black!30} & \cellcolor{black!50} 1 & \cellcolor{red!70} 1 \\ 
\cellcolor{black!30} & \cellcolor{black!30} & \cellcolor{black!30} & \cellcolor{black!50} 0 & \cellcolor{red!70} 1 \\ 
\cellcolor{black!30} & \cellcolor{black!30} & \cellcolor{black!30} & \cellcolor{black!50} 1 & \cellcolor{red!70} 1 \\ 
\cellcolor{black!30} & \cellcolor{black!30} & \cellcolor{black!30} & \cellcolor{black!50} 0 & \cellcolor{red!70} 1 \\ 
\cellcolor{black!30} & \cellcolor{black!30} & \cellcolor{black!30} & \cellcolor{black!50} 0 & \cellcolor{red!70} 1 \\ 
\hline
\end{tabular}
}
\hspace{2pt} \text{ to predict } \mathbf y_1
\label{eq:truevspredtrue}
\end{equation}

For an unlabeled observation $\mathbf x^{(l)}$, only the three predictor variables $x^{(l)}_1,\hdots,x^{(l)}_3$ are available 
at prediction time. However, the classifier $C_1^\text{true}$ needs a 4-dimensional observation $\left(\mathbf x^{(l)}, y^{(l)}_3\right)$ as input. The input $y^{(l)}_3$ therefore 
needs to be predicted first. A new \textit{level-1} classifier $C_3^\text{lvl1}$, which is trained on the set $D' =\cup_{i=1}^6\left\{\left(\mathbf x^{(i)},y^{(i)}_3\right)\right\}$, will make 
those predictions for $y^{(l)}_3$. The training task is:
\begin{equation}
\text{Train } C_3^\text{1vl1} \text{ on } \hspace{2pt}
D' \hspace{2pt} \hat = \hspace{2pt} 
\scalebox{.55}{
\begin{tabular}{|lll|@{}|@{}|@{}|@{}|@{}|@{}|@{}|@{}|l|}
\hline
$\mathbf x_1$ & $\mathbf x_2$ & $\mathbf x_3$ & $\mathbf y_3$ \\ 
\hline
\cellcolor{black!30} & \cellcolor{black!30} & \cellcolor{black!30} & \cellcolor{red!70} 1 \\ 
\cellcolor{black!30} & \cellcolor{black!30} & \cellcolor{black!30} & \cellcolor{red!70} 1 \\ 
\cellcolor{black!30} & \cellcolor{black!30} & \cellcolor{black!30}  & \cellcolor{red!70} 0 \\ 
\cellcolor{black!30} & \cellcolor{black!30} & \cellcolor{black!30}  & \cellcolor{red!70} 1 \\ 
\cellcolor{black!30} & \cellcolor{black!30} & \cellcolor{black!30} &  \cellcolor{red!70} 0 \\ 
\cellcolor{black!30} & \cellcolor{black!30} & \cellcolor{black!30} &  \cellcolor{red!70} 0 \\ 
\hline
\end{tabular}
}
\hspace{2pt} \text{ to predict } \mathbf y_3
\label{eq:predictionphase}
\end{equation}
Therefore, for a new observation $\mathbf x^{(l)}$, the predicted label $\hat y^{(l)}_3$ is obtained by using $C_3^\text{lvl1}$ on $\mathbf x^{(l)}$. The final prediction for
$y^{(l)}_1$ is then obtained by using $C_1^\text{true}$ on $\left(\mathbf x^{(l)},\hat y^{(l)}_3\right)$.

The alternative to (\ref{eq:truevspredtrue}) would be to use predicted labels $\mathbf{\hat{y}_3}$ instead of true labels $\mathbf y_3$. These labels should be 
produced by means of an out-of-sample prediction procedure \citep{Senge2013}. This can be done by an internal leave-one-out cross-validation procedure, 
which can of course be computationally intensive. Because of this, coarser resampling strategies can be used. As an example, an 
internal $2$-fold cross-validation will be shown here. Again, let $D' = \cup_{i=1}^{6} \left\{\left(\mathbf x^{(i)},y^{(i)}_3\right)\right\}$ be the set of all predictor variables with $\mathbf y_3$ as target variable.
Using 2-fold cross-validation, the dataset $D'$ is split into two parts $D'_1= \cup_{i=1}^{3} \left\{\left(\mathbf x^{(i)},y^{(i)}_3\right)\right\}$ and $D'_2= \cup_{i=4}^{6} \left\{\left(\mathbf x^{(i)},y^{(i)}_3\right)\right\}$:
\begin{equation}
\scalebox{.55}{
\begin{tabular}{|lcl|@{}|@{}|@{}|@{}|@{}|@{}|@{}|@{}|l|}
\hline
$\mathbf x_1$ & $\mathbf x_2$ & $\mathbf x_3$ & $\mathbf y_3$ \\ 
\hline
\cellcolor{black!30} & \cellcolor{black!30} & \cellcolor{black!30} &  \cellcolor{red!70} 1 \\ 
\cellcolor{black!30} & \cellcolor{black!30} \large{$D'_1$} & \cellcolor{black!30} & \cellcolor{red!70} 1 \\ 
\cellcolor{black!30} & \cellcolor{black!30} & \cellcolor{black!30} & \cellcolor{red!70} 0 \\ 
\specialrule{.3em}{0em}{0em} 
\cellcolor{black!30} & \cellcolor{black!30} & \cellcolor{black!30} & \cellcolor{red!70} 1 \\ 
\cellcolor{black!30} & \cellcolor{black!30} \large{$D'_2$} & \cellcolor{black!30} &\cellcolor{red!70} 0 \\ 
\cellcolor{black!30} & \cellcolor{black!30} & \cellcolor{black!30} &  \cellcolor{red!70} 0 \\ 
\hline
\end{tabular}
}
\label{eq:level1predtruevspred2}
\end{equation}

Two classifiers $C_{D'_1}$ and $C_{D'_2}$ are then trained on $D'_1$ and $D'_2$, respectively, for the prediction of $\mathbf y_3$:
\begin{align*}
\text{Train } C_{D'_1} \text{ on } \hspace{2pt}
\scalebox{.55}{
\begin{tabular}{|lcl|@{}|@{}|@{}|@{}|@{}|@{}|@{}|@{}|l|}
\hline
$\mathbf x_1$ & $\mathbf x_2$ & $\mathbf x_3$ & $\mathbf y_3$ \\ 
\hline
\cellcolor{black!30} & \cellcolor{black!30} & \cellcolor{black!30} & \cellcolor{red!70} 1 \\ 
\cellcolor{black!30} & \cellcolor{black!30} \large{$D'_1$} & \cellcolor{black!30} & \cellcolor{red!70} 1 \\ 
\cellcolor{black!30} & \cellcolor{black!30} & \cellcolor{black!30}  & \cellcolor{red!70} 0 \\
\hline
\end{tabular}
}
\hspace{2pt} \text{ to predict } \mathbf y_3, \quad
\text{Train } C_{D'_2} \text{ on } \hspace{2pt}
\scalebox{.55}{
\begin{tabular}{|lcl|@{}|@{}|@{}|@{}|@{}|@{}|@{}|@{}|l|}
\hline
$\mathbf x_1$ & $\mathbf x_2$ & $\mathbf x_3$ &  $\mathbf y_3$ \\ 
\hline
\cellcolor{black!30} & \cellcolor{black!30} & \cellcolor{black!30} & \cellcolor{red!70} 1 \\ 
\cellcolor{black!30} & \cellcolor{black!30} \large{$D'_2$}& \cellcolor{black!30} & \cellcolor{red!70} 0 \\ 
\cellcolor{black!30} & \cellcolor{black!30} & \cellcolor{black!30} & \cellcolor{red!70} 0 \\ 
\hline
\end{tabular}
}
\hspace{2pt} \text{ to predict } \mathbf y_3
\end{align*}

\newpage

Following the cross-validation paradigm, $D'_1$ is used as test set for the classifier $C_{D'_2}$, 
and $D'_2$ is used as a test set for $C_{D'_1}$:
\begin{align*}
C_{D'_2} : 
\scalebox{.55}{
\begin{tabular}{|lcl|}
\hline
$\mathbf x_1$ & $\mathbf x_2$ & $\mathbf x_3$  \\ 
\hline
\cellcolor{black!30} & \cellcolor{black!30} & \cellcolor{black!30}  \\ 
\cellcolor{black!30} & \cellcolor{black!30} \large{$D'_1$} & \cellcolor{black!30}  \\ 
\cellcolor{black!30} & \cellcolor{black!30} & \cellcolor{black!30}  \\ 
\hline
\end{tabular}
}
\mapsto  \scalebox{.55}{
\begin{tabular}{|l|}
\hline
$ \mathbf{\hat{y}_3}$  \\ 
\hline
\cellcolor{black!50} 1 \\ 
\cellcolor{black!50} 0  \\ 
\cellcolor{black!50} 0 \\
\hline
\end{tabular}
}, \quad
C_{D'_1} : 
\scalebox{.55}{
\begin{tabular}{|lcl|}
\hline
$\mathbf x_1$ & $\mathbf x_2$ & $\mathbf x_3$ \\ 
\hline
\cellcolor{black!30} & \cellcolor{black!30} & \cellcolor{black!30}   \\ 
\cellcolor{black!30} & \cellcolor{black!30} \large{$D'_2$} & \cellcolor{black!30}   \\ 
\cellcolor{black!30} & \cellcolor{black!30} & \cellcolor{black!30}   \\ 
\hline
\end{tabular}
}
\mapsto  \scalebox{.55}{
\begin{tabular}{|l|}
\hline
$\mathbf{\hat{y}_3}$  \\ 
\hline
\cellcolor{black!50} 0 \\ 
\cellcolor{black!50} 0  \\ 
\cellcolor{black!50} 1 \\
\hline
\end{tabular}
}
\end{align*}

These predictions are merged for the final predicted label $\mathbf{\hat{y}_3}$, which is used to augment the feature space.
The classifier $C_1^\text{pred}$ is then trained on that augmented feature space:
\begin{equation}
\text{Train } C_1^\text{pred} \text{ on } \hspace{2pt}
\scalebox{.55}{
\begin{tabular}{|llll|@{}|@{}|@{}|@{}|@{}|@{}|@{}|@{}|l|}
\hline
$\mathbf x_1$ & $\mathbf x_2$ & $\mathbf x_3$ & $\mathbf{\hat{y}_3}$ & $\mathbf y_1$ \\ 
\hline
\cellcolor{black!30} & \cellcolor{black!30} & \cellcolor{black!30} & \cellcolor{black!50} 1 & \cellcolor{red!70} 0 \\ 
\cellcolor{black!30} & \cellcolor{black!30} & \cellcolor{black!30} & \cellcolor{black!50} 0 & \cellcolor{red!70} 1 \\ 
\cellcolor{black!30} & \cellcolor{black!30} & \cellcolor{black!30} & \cellcolor{black!50} 0 & \cellcolor{red!70} 1 \\ 
\cellcolor{black!30} & \cellcolor{black!30} & \cellcolor{black!30} & \cellcolor{black!50} 0 & \cellcolor{red!70} 1 \\ 
\cellcolor{black!30} & \cellcolor{black!30} & \cellcolor{black!30} & \cellcolor{black!50} 0 & \cellcolor{red!70} 1 \\ 
\cellcolor{black!30} & \cellcolor{black!30} & \cellcolor{black!30} & \cellcolor{black!50} 1 & \cellcolor{red!70} 1 \\ 
\hline
\end{tabular}
}
\hspace{2pt} \text{ to predict } \mathbf y_1
\end{equation}

The prediction phase is completely analogous to (\ref{eq:predictionphase}). It is worthwhile to mention that the level-1 classifier $C^\text{lvl1}_{3}$,  
which will be used to obtain predictions $\mathbf{\hat{y}_3}$ at prediction time, is trained on the whole set $D' = D'_1 \cup D'_2$, following \citet{Simon2007}.

\subsubsection{Full vs. partial conditioning}
Recall the set of all labels $Z = \lbrace z_1, \hdots, z_m \rbrace$. The prediction of a label $z_k$ can either be conditioned on all remaining 
labels $\lbrace z_1,\hdots,z_{k-1},z_{k+1},\hdots,z_m\rbrace$ (\textit{full conditioning}) or just on a subset of labels (\textit{partial conditioning}). 
The only method for partial conditioning, which is examined in this paper, is the chaining method. Here, labels $z_k$ are conditioned on all previous 
labels $\lbrace z_1, \hdots, z_{k-1} \rbrace$ for all $k = 1, \hdots, m$. This sequential structure is motivated by the product rule of probability \citep{Montanes2014}:
\begin{equation}
P\left(\mathbf y^{(i)} \middle| \mathbf x^{(i)} \right) = \prod_{k = 1}^{m} P\left(y^{(i)}_k\middle|\mathbf x^{(i)},y^{(i)}_1,\hdots,y^{(i)}_{k-1}\right)
\label{eq:conditionalprob}
\end{equation}
Methods that make use of this chaining structure are e.g., \textit{classifier chains} or \textit{nested stacking} (these methods will be discussed 
further below). 

To sum up the discussions above: there are four ways in modeling label dependencies through conditioning labels $z_{k}$ on other labels $z_{\ell}$, $k\neq \ell$. They can be distinguished 
by the subset of labels, which are used for conditioning, and by the use of predicted or real labels in the training step. In \\ Table \ref{tab:4methods} we show the four methods, 
which implement these ideas and describe them consequently.
\begin{table}[ht]
\centering
\begin{tabular}{rll}
\toprule
& True labels & Pred. labels \\ 
\midrule
Partial cond. & Classifier chains & Nested stacking \\ 
Full cond. & Dependent binary relevance & Stacking \\ 
\bottomrule
\end{tabular}
\caption{Distinctions in modeling label dependence and models}
\label{tab:4methods}
\end{table}

\subsection{Classifier chains}
\label{sec:CC}
The classifier chains (CC) method implements the idea of using partial conditioning together with the true label information. It was first introduced by \citet{Read2011}. 
CC selects an order  on the set of labels $\lbrace z_1, \hdots,z_m \rbrace$, which can be formally written as a bijective function (permutation):
\begin{equation}
\tau : \lbrace 1,\hdots,m\rbrace \longrightarrow \lbrace 1, \hdots, m \rbrace
\end{equation}
Labels will be chained along this order $\tau$:
\begin{equation}
z_{\tau(1)} \rightarrow z_{\tau(2)} \rightarrow \hdots \rightarrow z_{\tau(m)}
\end{equation}
However, for this paper the permutation shall be $\tau = id$ (only for simplicity reasons). The labels therefore follow the order $z_1 \rightarrow z_2 \rightarrow \hdots \rightarrow z_m$. 
In a similar fashion to the binary relevance (BR) method, CC trains $m$ binary classifiers $C_k$, which are responsible for predicting their corresponding label $z_k$, $k = 1, \hdots, m$. 
The classifiers $C_k$ are of the form
\begin{equation}
C_k : \mathcal X \times \lbrace 0,1 \rbrace^{k-1} \longrightarrow \lbrace 0,1\rbrace,
\end{equation}
where $\lbrace 0,1 \rbrace^0 := \emptyset$. For a classifier $C_k$ the feature space is augmented by the true label information of all previous labels $z_1, z_2, \hdots, z_{k-1}$. Hence, 
the training data of $C_k$ consists of all observations $\left(\left(\mathbf x^{(i)}, y_1^{(i)},  y_2^{(i)},\hdots,y_{k-1}^{(i)}\right),  y_k^{(i)}\right)$, $i=1,\hdots,n$, with the target $y_k^{(i)}$. 
In the example from above, this would look like:
\begin{align}
\text{Train } C_1 \text{ on } \hspace{2pt}\scalebox{.55}{
\begin{tabular}{|lll|@{}|@{}|@{}|@{}|@{}|@{}|@{}|@{}|l|}
\hline
$\mathbf x_1$ & $\mathbf x_2$ & $\mathbf x_3$ & $\mathbf y_1$ \\ 
\hline
\cellcolor{black!30} & \cellcolor{black!30} & \cellcolor{black!30} & \cellcolor{red!70} 0 \\ 
\cellcolor{black!30} & \cellcolor{black!30} & \cellcolor{black!30} & \cellcolor{red!70} 1 \\ 
\cellcolor{black!30} & \cellcolor{black!30} & \cellcolor{black!30} & \cellcolor{red!70} 1 \\ 
\cellcolor{black!30} & \cellcolor{black!30} & \cellcolor{black!30} & \cellcolor{red!70} 1\\ 
\cellcolor{black!30} & \cellcolor{black!30} & \cellcolor{black!30} & \cellcolor{red!70} 1\\ 
\cellcolor{black!30} & \cellcolor{black!30} & \cellcolor{black!30} & \cellcolor{red!70} 1\\ 
\hline
\end{tabular}
} \hspace{10pt}
\text{Train } C_2 \text{ on } \hspace{2pt}\scalebox{.55}{
\begin{tabular}{|llll|@{}|@{}|@{}|@{}|@{}|@{}|@{}|@{}|l|}
\hline
$\mathbf x_1$ & $\mathbf x_2$ & $\mathbf x_3$ & $\mathbf y_1$ & $\mathbf y_2$ \\ 
\hline
\cellcolor{black!30} & \cellcolor{black!30} & \cellcolor{black!30} & \cellcolor{black!50} 0 & \cellcolor{red!70} 0\\ 
\cellcolor{black!30} & \cellcolor{black!30} & \cellcolor{black!30} & \cellcolor{black!50} 1 & \cellcolor{red!70} 0\\ 
\cellcolor{black!30} & \cellcolor{black!30} & \cellcolor{black!30} & \cellcolor{black!50} 1 & \cellcolor{red!70} 1\\ 
\cellcolor{black!30} & \cellcolor{black!30} & \cellcolor{black!30} & \cellcolor{black!50} 1 & \cellcolor{red!70} 1\\ 
\cellcolor{black!30} & \cellcolor{black!30} & \cellcolor{black!30} & \cellcolor{black!50} 1 & \cellcolor{red!70} 1\\ 
\cellcolor{black!30} & \cellcolor{black!30} & \cellcolor{black!30} & \cellcolor{black!50} 1 & \cellcolor{red!70} 1\\ 
\hline
\end{tabular}
} \hspace{10pt}
\text{Train } C_3 \text{ on } \hspace{2pt}\scalebox{.55}{
\begin{tabular}{|lllll|@{}|@{}|@{}|@{}|@{}|@{}|@{}|@{}|l|}
\hline
$\mathbf x_1$ & $\mathbf x_2$ & $\mathbf x_3$ & $\mathbf y_1$ & $\mathbf y_2$ & $\mathbf y_3$ \\ 
\hline
\cellcolor{black!30} & \cellcolor{black!30} & \cellcolor{black!30} & \cellcolor{black!50} 0 & \cellcolor{black!50} 0 & \cellcolor{red!70} 1\\ 
\cellcolor{black!30} & \cellcolor{black!30} & \cellcolor{black!30} & \cellcolor{black!50} 1 & \cellcolor{black!50} 0 & \cellcolor{red!70} 1\\ 
\cellcolor{black!30} & \cellcolor{black!30} & \cellcolor{black!30} & \cellcolor{black!50} 1 & \cellcolor{black!50} 1 & \cellcolor{red!70} 0\\ 
\cellcolor{black!30} & \cellcolor{black!30} & \cellcolor{black!30} & \cellcolor{black!50} 1 & \cellcolor{black!50} 1 & \cellcolor{red!70} 1\\ 
\cellcolor{black!30} & \cellcolor{black!30} & \cellcolor{black!30} & \cellcolor{black!50} 1 & \cellcolor{black!50} 1 & \cellcolor{red!70} 0\\ 
\cellcolor{black!30} & \cellcolor{black!30} & \cellcolor{black!30} & \cellcolor{black!50} 1 & \cellcolor{black!50} 1 & \cellcolor{red!70} 0\\ 
\hline
\end{tabular}
}
\end{align}	

At prediction time, when an unlabeled observation $\mathbf x^{(l)}$ is labeled, a prediction $\left(\hat y_1^{(l)}, \hdots, \hat y_m^{(l)}\right)$ is obtained by successively predicting the 
labels along the chaining order: 
\begin{align}
\begin{split}
\hat y_1^{(l)} &= C_1 \left(\mathbf x^{(l)}\right) \\
\hat y_2^{(l)} &= C_2 \left(\mathbf x^{(l)},\hat y_1^{(l)}\right)\\
\vdots \\
\hat y_m^{(l)} &= C_m \left(\mathbf x^{(l)},\hat y_1^{(l)}, \hat y_2^{(l)}, \hdots, \hat y_{m-1}^{(l)}\right)
\end{split}
\label{eq:predCC}
\end{align}

The authors of \citet{Senge2013} summarize several factors, which have an impact on the performance of CC:
\begin{itemize}
\item \textit{The length of the chain.} A high number ($k-1$) of preceding classifiers in the chain comes with a high potential level of feature noise for the classifier $C_k$.
One may assume that the probability of a mistake will increase with the level of feature noise in the input space. Then the probability of a mistake will 
be reinforced along the chain, due to the recursive structure of CC.
\item \textit{The order of the chain.} Some labels may be more difficult to predict than others. The order of a chain can therefore be important for the performance. 
It can be advantageous to put simple to predict labels in the beginning and harder to predict labels more towards the end of the chain. 
Some heuristics for finding an optimal chain ordering have been proposed in \citet{Silva2014, Read2013}. Alternatively \citet{Read2011} developed 
an ensemble of classifier chains, which builds many randomly ordered CC-classifiers and put them on a voting scheme for a prediction. However, these methods are 
not subject of this article.
\item \textit{The dependency among labels.} For an improvement of performance through chaining, there should be a dependence among labels, CC cannot gain in case of label independence. 
However, CC is also only likely to lose if the binary classifiers $C_k$ cannot ignore the added features $\mathbf y_1, \hdots, \mathbf y_{k-1}$.
\end{itemize}

\subsection{Nested stacking}
\label{sec:NST}
The nested stacking method (NST), first proposed in \citet{Senge2013}, implements the idea
of using partial conditioning together with predicted label information. NST mimicks the chaining structure of CC, but does not use real label information during training. 
Like in CC the chaining order shall be $\tau = id$ , again for simplicity reasons. 
CC uses real label information $\mathbf y_k$ during training and predicted labels $\mathbf{\hat{y}_k}$ at prediction time. However, unless the binary classifiers are perfect, it is likely 
that $\mathbf y_k$ and $\mathbf{\hat{y}_k}$ do not follow the same distribution. Hence, the key assumption of supervised learning, namely that the training data should be representative for 
the test data, is violated by CC.
Nested stacking tries to overcome this issue by using predicted labels $\mathbf{\hat{y}_k}$ instead of true labels $\mathbf y_k$.

NST trains $m$ binary classifiers $C_k$ on $D_k := \cup_{i=1}^{n} \left\{\left(\left(\mathbf x^{(i)},\hat y_1^{(i)},\hdots,\hat y_{k-1}^{(i)}\right),y_k^{(i)}\right)\right\}$, for all $k = 1, \hdots, m$. The predicted labels should be 
obtained by an internal out-of-sample method \citep{Senge2013}.
How these predictions are obtained was already explained in the \textbf{True vs. Predicted Label Information} chapter. 
The prediction phase is completely analogous to (\ref{eq:predCC}). 

The training procedure is visualized in the following with 2-fold cross-validation as an internal out-of-sample method:
\begin{align}
\text{Train } C_1 \text{ on } \hspace{2pt}\scalebox{.55}{
\begin{tabular}{|lll|@{}|@{}|@{}|@{}|@{}|@{}|@{}|@{}|l|}
\hline
$\mathbf x_1$ & $\mathbf x_2$ & $\mathbf x_3$ & $\mathbf y_1$ \\ 
\hline
\cellcolor{black!30} & \cellcolor{black!30} & \cellcolor{black!30} & \cellcolor{red!70} 0 \\ 
\cellcolor{black!30} & \cellcolor{black!30} & \cellcolor{black!30} & \cellcolor{red!70} 1 \\ 
\cellcolor{black!30} & \cellcolor{black!30} & \cellcolor{black!30} & \cellcolor{red!70} 1 \\ 
\cellcolor{black!30} & \cellcolor{black!30} & \cellcolor{black!30} & \cellcolor{red!70} 1\\ 
\cellcolor{black!30} & \cellcolor{black!30} & \cellcolor{black!30} & \cellcolor{red!70} 1\\ 
\cellcolor{black!30} & \cellcolor{black!30} & \cellcolor{black!30} & \cellcolor{red!70} 1\\ 
\hline
\end{tabular}
} \hspace{50pt}
\text{Use 2-fold CV on} \hspace{2pt}\scalebox{.55}{
\begin{tabular}{|lll|@{}|@{}|@{}|@{}|@{}|@{}|@{}|@{}|l|}
\hline
$\mathbf x_1$ & $\mathbf x_2$ & $\mathbf x_3$ & $\mathbf y_1$ \\ 
\hline
\cellcolor{black!30} & \cellcolor{black!30} & \cellcolor{black!30} & \cellcolor{red!70} 0 \\ 
\cellcolor{black!30} & \cellcolor{black!30} & \cellcolor{black!30} & \cellcolor{red!70} 1 \\ 
\cellcolor{black!30} & \cellcolor{black!30} & \cellcolor{black!30} & \cellcolor{red!70} 1 \\ 
\cellcolor{black!30} & \cellcolor{black!30} & \cellcolor{black!30} & \cellcolor{red!70} 1\\ 
\cellcolor{black!30} & \cellcolor{black!30} & \cellcolor{black!30} & \cellcolor{red!70} 1\\ 
\cellcolor{black!30} & \cellcolor{black!30} & \cellcolor{black!30} & \cellcolor{red!70} 1\\ 
\hline
\end{tabular}
} \hspace{2pt} \text{to obtain} \hspace{2pt} \scalebox{.55}{
\begin{tabular}{|l|}
\hline
$\hat{ \mathbf y_1}$  \\ 
\hline
\cellcolor{black!50} 1 \\ 
\cellcolor{black!50} 1  \\ 
\cellcolor{black!50} 1  \\ 
\cellcolor{black!50} 1  \\ 
\cellcolor{black!50} 0  \\ 
\cellcolor{black!50} 1
\end{tabular}
}
\end{align}
\begin{align}
\text{Train } C_2 \text{ on } \hspace{2pt}
\scalebox{.55}{
\begin{tabular}{|llll|@{}|@{}|@{}|@{}|@{}|@{}|@{}|@{}|l|}
\hline
$\mathbf x_1$ & $\mathbf x_2$ & $\mathbf x_3$ & $\mathbf{\hat{y}_1}$ & $\mathbf y_2$ \\ 
\hline
\cellcolor{black!30} & \cellcolor{black!30} & \cellcolor{black!30} & \cellcolor{black!50} 1 & \cellcolor{red!70} 0 \\ 
\cellcolor{black!30} & \cellcolor{black!30} & \cellcolor{black!30} & \cellcolor{black!50} 1 & \cellcolor{red!70} 0 \\ 
\cellcolor{black!30} & \cellcolor{black!30} & \cellcolor{black!30} & \cellcolor{black!50} 1 & \cellcolor{red!70} 1 \\ 
\cellcolor{black!30} & \cellcolor{black!30} & \cellcolor{black!30} & \cellcolor{black!50} 1 & \cellcolor{red!70} 1 \\ 
\cellcolor{black!30} & \cellcolor{black!30} & \cellcolor{black!30} & \cellcolor{black!50} 0 & \cellcolor{red!70} 1 \\ 
\cellcolor{black!30} & \cellcolor{black!30} & \cellcolor{black!30} & \cellcolor{black!50} 1 & \cellcolor{red!70} 1 \\ 
\hline
\end{tabular}
}
\hspace{30pt}
\text{Use 2-fold CV on} \hspace{2pt}\scalebox{.55}{
\begin{tabular}{|llll|@{}|@{}|@{}|@{}|@{}|@{}|@{}|@{}|l|}
\hline
$\mathbf x_1$ & $\mathbf x_2$ & $\mathbf x_3$ & $\mathbf{\hat{y}_1}$ & $\mathbf y_2$ \\ 
\hline
\cellcolor{black!30} & \cellcolor{black!30} & \cellcolor{black!30} & \cellcolor{black!50} 1 & \cellcolor{red!70} 0 \\ 
\cellcolor{black!30} & \cellcolor{black!30} & \cellcolor{black!30} & \cellcolor{black!50} 1 & \cellcolor{red!70} 0 \\ 
\cellcolor{black!30} & \cellcolor{black!30} & \cellcolor{black!30} & \cellcolor{black!50} 1 & \cellcolor{red!70} 1 \\ 
\cellcolor{black!30} & \cellcolor{black!30} & \cellcolor{black!30} & \cellcolor{black!50} 1 & \cellcolor{red!70} 1 \\ 
\cellcolor{black!30} & \cellcolor{black!30} & \cellcolor{black!30} & \cellcolor{black!50} 0 & \cellcolor{red!70} 1 \\ 
\cellcolor{black!30} & \cellcolor{black!30} & \cellcolor{black!30} & \cellcolor{black!50} 1 & \cellcolor{red!70} 1 \\ 
\hline
\end{tabular}
} \hspace{2pt} \text{to obtain} \hspace{2pt} \scalebox{.55}{
\begin{tabular}{|l|}
\hline
$\mathbf{\hat{y}_2}$  \\ 
\hline
\cellcolor{black!50} 1 \\ 
\cellcolor{black!50} 1  \\ 
\cellcolor{black!50} 1  \\ 
\cellcolor{black!50} 0  \\ 
\cellcolor{black!50} 1  \\ 
\cellcolor{black!50} 0
\end{tabular}
}
\end{align}
\begin{align}
\text{Train } C_3 \text{ on } \hspace{2pt}
\scalebox{.55}{
\begin{tabular}{|lllll|@{}|@{}|@{}|@{}|@{}|@{}|@{}|@{}|l|}
\hline
$\mathbf x_1$ & $\mathbf x_2$ & $\mathbf x_3$ & $\mathbf{\hat{y}_1}$ & $\mathbf{\hat{y}_2}$ & $\mathbf y_3$ \\ 
\hline
\cellcolor{black!30} & \cellcolor{black!30} & \cellcolor{black!30} & \cellcolor{black!50} 1 & \cellcolor{black!50} 1 & \cellcolor{red!70} 1 \\ 
\cellcolor{black!30} & \cellcolor{black!30} & \cellcolor{black!30} & \cellcolor{black!50} 1 & \cellcolor{black!50} 1 & \cellcolor{red!70} 1 \\ 
\cellcolor{black!30} & \cellcolor{black!30} & \cellcolor{black!30} & \cellcolor{black!50} 1  & \cellcolor{black!50} 1 & \cellcolor{red!70} 0 \\ 
\cellcolor{black!30} & \cellcolor{black!30} & \cellcolor{black!30} & \cellcolor{black!50} 1 & \cellcolor{black!50} 0 & \cellcolor{red!70} 1 \\ 
\cellcolor{black!30} & \cellcolor{black!30} & \cellcolor{black!30} & \cellcolor{black!50} 0 & \cellcolor{black!50} 1 & \cellcolor{red!70} 0 \\ 
\cellcolor{black!30} & \cellcolor{black!30} & \cellcolor{black!30} & \cellcolor{black!50} 1 & \cellcolor{black!50} 0 & \cellcolor{red!70} 0 \\ 
\hline
\end{tabular}
} \hspace{180pt}
\end{align}

The factors which impact the performance of CC (i.e., length and order of the chain, and the dependency among labels), also impact NST, since NST mimicks the chaining 
method of CC. 

\subsection{Dependent binary relevance}
\label{sec:DBR}
The dependent binary relevance method (DBR) implements the idea of using full conditioning together with the true label information. DBR is built on two main hypotheses 
\citep{Montanes2014}: 
\begin{itemize}
\item[(i)] Taking conditional label dependencies into account is important for performing well in multilabel classification tasks. 
\item[(ii)] Modeling and learning these label dependencies in an overcomplete way (take all other labels for modeling) may further improve model performance.
\end{itemize}
The first assumption is the main prerequisite for research in multilabel classification. It has been shown theoretically that simple binary relevance classifiers cannot 
achieve optimal performance for specific multilabel loss functions \citep{Montanes2014}.
The second assumption, however, is harder to justify theoretically. Nonetheless, the practical usefulness of learning in an overcomplete way has been shown in many branches of 
(classical) single-label classification  (e.g., ensemble methods \citep{Dietterich2000}). 

Formally, DBR trains $m$ binary classifiers $C_1, \hdots, C_m$ (as many classifiers as labels) on the corresponding training data
\begin{equation}
D_k = \cup_{i=1}^n \left\{\left(\left(\mathbf x^{(i)}, y_1^{(i)}, \hdots,  y_{k-1}^{(i)}, y_{k+1}^{(i)},\hdots,y_m^{(i)}\right), y_k^{(i)}\right)\right\}, 
\label{eq:trainingdataDBR}
\end{equation}
$k=1,\hdots,m$. Thus, each classifier $C_k$ is of the form
$$
C_k : \mathcal X \times \lbrace 0,1 \rbrace^{m-1} \longrightarrow \lbrace 0,1  \rbrace.
$$
Hence, for each classifier $C_k$ the true label information of all labels except $\mathbf y_k$  is used as augmented features. Again, here is a visualization with the example from above:
\begin{align}
\text{Train } C_1 \text{ on} \hspace{1pt}\scalebox{.55}{
\begin{tabular}{|lllll|@{}|@{}|@{}|@{}|@{}|@{}|@{}|@{}|l|}
\hline
$\mathbf x_1$ & $\mathbf x_2$ & $\mathbf x_3$ & $\mathbf y_2$ & $\mathbf y_3$ & $\mathbf y_1$ \\ 
\hline
\cellcolor{black!30} & \cellcolor{black!30} & \cellcolor{black!30} & \cellcolor{black!50} 0 & \cellcolor{black!50} 1 & \cellcolor{red!70} 0\\ 
\cellcolor{black!30} & \cellcolor{black!30} & \cellcolor{black!30} & \cellcolor{black!50} 0  & \cellcolor{black!50} 1 & \cellcolor{red!70} 1\\ 
\cellcolor{black!30} & \cellcolor{black!30} & \cellcolor{black!30} & \cellcolor{black!50} 1 & \cellcolor{black!50} 0 & \cellcolor{red!70} 1\\ 
\cellcolor{black!30} & \cellcolor{black!30} & \cellcolor{black!30} & \cellcolor{black!50} 1 & \cellcolor{black!50} 1 & \cellcolor{red!70} 1\\ 
\cellcolor{black!30} & \cellcolor{black!30} & \cellcolor{black!30} & \cellcolor{black!50} 1 & \cellcolor{black!50} 0 & \cellcolor{red!70} 1\\ 
\cellcolor{black!30} & \cellcolor{black!30} & \cellcolor{black!30} & \cellcolor{black!50} 1 & \cellcolor{black!50} 0 & \cellcolor{red!70} 1\\ 
\hline
\end{tabular}
} \hspace{10pt}
\text{Train } C_2 \text{ on} \hspace{1pt}\scalebox{.55}{
\begin{tabular}{|lllll|@{}|@{}|@{}|@{}|@{}|@{}|@{}|@{}|l|}
\hline
$\mathbf x_1$ & $\mathbf x_2$ & $\mathbf x_3$ & $\mathbf y_1$ & $\mathbf y_3$ & $\mathbf y_2$ \\ 
\hline
\cellcolor{black!30} & \cellcolor{black!30} & \cellcolor{black!30} & \cellcolor{black!50} 0 & \cellcolor{black!50} 1 & \cellcolor{red!70} 0\\ 
\cellcolor{black!30} & \cellcolor{black!30} & \cellcolor{black!30} & \cellcolor{black!50} 1  & \cellcolor{black!50} 1 & \cellcolor{red!70} 0\\ 
\cellcolor{black!30} & \cellcolor{black!30} & \cellcolor{black!30} & \cellcolor{black!50} 1 & \cellcolor{black!50} 0 & \cellcolor{red!70} 1\\ 
\cellcolor{black!30} & \cellcolor{black!30} & \cellcolor{black!30} & \cellcolor{black!50} 1 & \cellcolor{black!50} 1 & \cellcolor{red!70} 1\\ 
\cellcolor{black!30} & \cellcolor{black!30} & \cellcolor{black!30} & \cellcolor{black!50} 1 & \cellcolor{black!50} 0 & \cellcolor{red!70} 1\\ 
\cellcolor{black!30} & \cellcolor{black!30} & \cellcolor{black!30} & \cellcolor{black!50} 1 & \cellcolor{black!50} 0 & \cellcolor{red!70} 1\\ 
\hline
\end{tabular}
} \hspace{10pt}
\text{Train } C_3 \text{ on} \hspace{1pt}\scalebox{.55}{
\begin{tabular}{|lllll|@{}|@{}|@{}|@{}|@{}|@{}|@{}|@{}|l|}
\hline
$\mathbf x_1$ & $\mathbf x_2$ & $\mathbf x_3$ & $\mathbf y_1$ & $\mathbf y_2$ & $\mathbf y_3$ \\ 
\hline
\cellcolor{black!30} & \cellcolor{black!30} & \cellcolor{black!30} & \cellcolor{black!50} 0 & \cellcolor{black!50} 0 & \cellcolor{red!70} 1\\ 
\cellcolor{black!30} & \cellcolor{black!30} & \cellcolor{black!30} & \cellcolor{black!50} 1  & \cellcolor{black!50} 0 & \cellcolor{red!70} 1\\ 
\cellcolor{black!30} & \cellcolor{black!30} & \cellcolor{black!30} & \cellcolor{black!50} 1 & \cellcolor{black!50} 1 & \cellcolor{red!70} 0\\ 
\cellcolor{black!30} & \cellcolor{black!30} & \cellcolor{black!30} & \cellcolor{black!50} 1 & \cellcolor{black!50} 1 & \cellcolor{red!70} 1\\ 
\cellcolor{black!30} & \cellcolor{black!30} & \cellcolor{black!30} & \cellcolor{black!50} 1 & \cellcolor{black!50} 1 & \cellcolor{red!70} 0\\ 
\cellcolor{black!30} & \cellcolor{black!30} & \cellcolor{black!30} & \cellcolor{black!50} 1 & \cellcolor{black!50} 1 & \cellcolor{red!70} 0\\ 
\hline
\end{tabular}
} 
\end{align}

To make these classifiers applicable, when an unlabeled instance $\mathbf x^{(l)}$ needs to be labeled, the help of other multilabel classifiers is needed to produce predicted 
labels $\hat y^{(l)}_1, \hdots., \hat y^{(l)}_m$ as additional features. 
The classifiers, which produce predicted labels as additional features, are called \textit{base learners} \citep{Montanes2014}. Theoretically any multilabel classifier can 
be used as base learner. However, in this paper, the analysis is focused on BR as base learner only. The prediction of an unlabeled instance $\mathbf x^{(l)}$ formally works as follows:
\begin{itemize}
\item[(i)] First level: Produce predicted labels by using the BR base learner:
$$
C_{BR}\left(\mathbf x^{(l)}\right) = \left(\hat y^{(l)}_1, \hdots, \hat y^{(l)}_m\right)
$$

\newpage

\item[(ii)] Second level, which is also called meta level \citep{Montanes2014}: Produce final prediction $\hat{\hat {\mathbf y}}_k = \left(\hat{\hat y}^{(l)}_1, \hdots, \hat{\hat y}^{(l)}_m \right)$ by 
applying DBR classifiers $C_1, \hdots, C_m$:
\begin{align*}  
C_1\left(\mathbf x^{(l)},\hat y^{(l)}_2,\hdots,\hat y^{(l)}_m\right) &= \hat{\hat y}^{(l)}_1 \\
C_2\left(\mathbf x^{(l)}, \hat y^{(l)}_1, \hat y^{(l)}_3, \hdots, \hat y^{(l)}_m\right)&= \hat{\hat y}^{(l)}_2 \\
& \vdots  \\
C_m\left(\mathbf x^{(l)},\hat y^{(l)}_1,\hdots,\hat y^{(l)}_{m-1}\right) &= \hat{\hat y}^{(l)}_m 
\end{align*}
\end{itemize}

\subsection{Stacking}
\label{sec:STA}
Stacking (STA) implements the last variant of Table \ref{tab:4methods}, namely the use of full conditioning together with predicted label information. 
Stacking is short for \textit{stacked generalization} \citep{Wolpert1992} and was first proposed  in the multilabel context by \citet{godbole2004}. 
Like in classical stacking, for each label it takes predictions of several other learners that were trained in a first step to get a new learner to make predictions for the 
corresponding label. Both hypotheses on which DBR is built on also apply to STA, of course.  

STA trains $m$ classifiers $C_1, \hdots, C_m$ on the corresponding training data 
\begin{equation}
D_k = \cup_{i=1}^n \left\{\left(\left(\mathbf x^{(i)}, \hat y_1^{(i)}, \hdots, \hat y_m^{(i)}\right), y_k^{(i)}\right)\right\}, k=1,\hdots,m.
\label{eq:trainingdataSTA}
\end{equation}
The classifiers $C_k$, $k=1, \hdots, m$, are therefore of the following form:
$$
C_k : \mathcal X \times \lbrace 0 , 1 \rbrace^m \longrightarrow \lbrace 0 , 1 \rbrace
$$
Like in NST, the predicted labels should be obtained by an internal out-of-sample method \citep{Sill2009}. STA can be seen as the alternative to DBR using predicted labels 
(like NST is for CC). However, the classifiers $C_k, k=1,\hdots,m,$ are trained on all predicted labels $\mathbf{\hat{y}_1}, \hdots, \mathbf{\hat{y}_m}$ for the STA approach (in DBR the label $\mathbf y_k$ is 
left out of the augmented training set).

The training procedure is outlined in the following:
\begin{align}
\text{For i=1,2,3 use 2-fold CV on }
\scalebox{.55}{
\begin{tabular}{|lll|@{}|@{}|@{}|@{}|@{}|@{}|@{}|@{}|c|}
\hline
$\mathbf x_1$ & $\mathbf x_2$ & $\mathbf x_3$ &  $\mathbf y_k$ \\ 
\hline
\cellcolor{black!30} & \cellcolor{black!30} & \cellcolor{black!30} & \cellcolor{red!70} $y^{(1)}_k$ \\ 
\cellcolor{black!30} & \cellcolor{black!30} & \cellcolor{black!30} & \cellcolor{red!70} $y^{(2)}_k$ \\ 
\cellcolor{black!30} & \cellcolor{black!30} & \cellcolor{black!30} & \cellcolor{red!70} $y^{(3)}_k$ \\ 
\cellcolor{black!30} & \cellcolor{black!30} & \cellcolor{black!30} & \cellcolor{red!70} $y^{(4)}_k$ \\ 
\cellcolor{black!30} & \cellcolor{black!30} & \cellcolor{black!30} & \cellcolor{red!70} $y^{(5)}_k$ \\ 
\cellcolor{black!30} & \cellcolor{black!30} & \cellcolor{black!30} & \cellcolor{red!70} $y^{(6)}_k$ \\ 
\hline
\end{tabular}
} \text{ to obtain } \scalebox{.55}{
\begin{tabular}{|l|}
\hline
$\hat y_k$  \\ 
\hline
\cellcolor{black!50} $\hat y^{(1)}_k$ \\ 
\cellcolor{black!50} $\hat y^{(2)}_k$  \\ 
\cellcolor{black!50} $\hat y^{(3)}_k$  \\ 
\cellcolor{black!50} $\hat y^{(4)}_k$  \\ 
\cellcolor{black!50} $\hat y^{(5)}_k$  \\ 
\cellcolor{black!50} $\hat y^{(6)}_k$
\end{tabular}
}
\end{align}
\begin{align}
\text{For i=1,2,3 train } C_k \text{ on} \hspace{1pt}\scalebox{.55}{
\begin{tabular}{|lllccc|@{}|@{}|@{}|@{}|@{}|@{}|@{}|@{}|c|}
\hline
$\mathbf x_1$ & $\mathbf x_2$ & $\mathbf x_3$ & $\hat y_1$ & $\hat y_2$ & $\hat y_3$ & $y_k$ \\ 
\hline
\cellcolor{black!30} & \cellcolor{black!30} & \cellcolor{black!30} & \cellcolor{black!50} $\hat y^{(1)}_1$ & \cellcolor{black!50} $\hat y^{(1)}_2$ & \cellcolor{black!50} $\hat y^{(1)}_3$ & \cellcolor{red!70} $y^{(1)}_k$\\ 
\cellcolor{black!30} & \cellcolor{black!30} & \cellcolor{black!30}  & \cellcolor{black!50} $\hat y^{(2)}_1$ & \cellcolor{black!50} $\hat y^{(2)}_2$  & \cellcolor{black!50} $\hat y^{(2)}_3$  & \cellcolor{red!70} $y^{(2)}_k$\\ 
\cellcolor{black!30} & \cellcolor{black!30} & \cellcolor{black!30}  & \cellcolor{black!50} $\hat y^{(3)}_1$ & \cellcolor{black!50} $\hat y^{(3)}_2$ & \cellcolor{black!50} $\hat y^{(3)}_3$  & \cellcolor{red!70} $y^{(3)}_k$\\ 
\cellcolor{black!30} & \cellcolor{black!30} & \cellcolor{black!30} & \cellcolor{black!50} $\hat y^{(4)}_1$  & \cellcolor{black!50} $\hat y^{(4)}_2$ & \cellcolor{black!50} $\hat y^{(4)}_3$  & \cellcolor{red!70} $y^{(4)}_k$\\ 
\cellcolor{black!30} & \cellcolor{black!30} & \cellcolor{black!30}  & \cellcolor{black!50} $\hat y^{(5)}_1$ & \cellcolor{black!50} $\hat y^{(5)}_2$ & \cellcolor{black!50} $\hat y^{(5)}_3$  & \cellcolor{red!70} $y^{(5)}_k$\\ 
\cellcolor{black!30} & \cellcolor{black!30} & \cellcolor{black!30}  & \cellcolor{black!50} $\hat y^{(6)}_1$ & \cellcolor{black!50} $\hat y^{(6)}_2$ & \cellcolor{black!50} $\hat y^{(6)}_3$  & \cellcolor{red!70} $y^{(6)}_k$\\ 
\hline
\end{tabular}
} 
\end{align}

Like in DBR, STA depends on a BR base learner, to produce predicted labels as additional features. Again, the use of BR as a base learner is not mandatory, but it is the proposed 
method in \citet{godbole2004}.

The prediction of an unlabeled instance $\mathbf x^{(l)}$ works almost identically to the DBR case and is illustrated here:
\begin{itemize}
\item[(i)] First level. Produce predicted labels by using the BR base learner:
$$
C_{BR}\left(\mathbf x^{(l)}\right) = \left(\hat y^{(l)}_1, \hdots, \hat y^{(l)}_m\right)
$$
\item[(ii)] Meta level. Apply STA classifiers $C_1, \hdots, C_m$:
\begin{align*}  
C_1\left(\mathbf x^{(l)},\hat y^{(l)}_1,\hdots,\hat y^{(l)}_m\right) &= \hat{\hat y}^{(l)}_1 \\
& \vdots  \\
C_m\left(\mathbf x^{(l)},\hat y^{(l)}_1,\hdots,\hat y^{(l)}_m\right) &= \hat{\hat y}^{(l)}_m 
\end{align*}
\end{itemize}

\subsection{Multilabel performance measures}
\label{sec:multilabelperformance}
Analogously to multiclass classification there exist multilabel classification performance measures. 
Six multilabel performance measures can be evaluated in \pkg{mlr}. These are:  \textit{Subset} $0/1$ \textit{loss}, 
\textit{hamming loss}, \textit{accuracy}, \textit{precision}, \textit{recall} and F$_1$-\textit{index}.
Multilabel performance measures are defined on a per instance basis. The performance on a test set is the average over all instances.

Let $D_{\text{test}} = \left\lbrace\left(\mathbf x^{(1)},\mathbf y^{(1)}\right),\hdots,\left(\mathbf x^{(n)},\mathbf y^{(n)}\right) \right\rbrace$ be a test set with 
$\mathbf y^{(i)} = \left(y_1^{(i)},\hdots,y_m^{(i)}\right) \in \lbrace 0,1\rbrace^m$ for all $i = 1,\hdots,n$. Performance measures quantify how good a classifier $C$ predicts the labels $z_1, \hdots, z_n$.

\begin{itemize}
\item[(i)] The subset $0/1$ loss is used to see if the predicted labels $C(\mathbf x^{(i)}) = \left(\hat y_1^{(i)},\hdots, \hat y_m^{(i)}\right)$ are equal to the actual labels $\left(y_1^{(i)},\hdots,y_m^{(i)}\right)$:
$$
\text{subset}_{0/1} \left( C, \left(\mathbf x^{(i)},\mathbf y^{(i)}\right) \right) = \mathds{1}_{\left(\mathbf{y}^{(i)} \neq C(\mathbf x^{(i)})\right)}:=
\begin{cases}  
1 & \text{if } \mathbf y^{(i)} \neq C\left(\mathbf x^{(i)}\right)\\
0 & \text{if }\mathbf y^{(i)} = C\left(\mathbf x^{(i)}\right)
\end{cases}
$$
The subset $0/1$ loss of a classifier $C$ on a test set $D_{\text{test}}$ thus becomes:
$$
\text{subset}_{0/1} \left( C, D_{\text{test}} \right) = \frac{1}{n} \sum_{i = 1}^n \mathds{1}_{\mathbf{y}^{(i)} \neq C\left(\mathbf x^{(i)}\right)}
$$
The subset 0/1 loss can be interpreted as the analogon of the mean misclassification error in multiclass classifications. 
In the multilabel case it is a rather drastic measure because it treats a mistake on a single label  as a complete failure \citep{Senge2013}.
\item[(ii)] The hamming loss also takes into account observations where only some labels have been predicted correctly. 
It corresponds to the proportion of labels whose relevance is incorrectly predicted. For an instance 
$\left(\mathbf x^{(i)}, \mathbf y^{(i)}\right) = \left(\mathbf x^{(i)}, \left(y_1^{(i)},\hdots,y_m^{(i)}\right)\right)$ and a classifier $C\left(\mathbf x^{(i)}\right) = \left(\hat y_1^{(i)},\hdots, \hat y_m^{(i)}\right)$ this is defined as:
$$
\text{HammingLoss}\left(C, \left(\mathbf x^{(i)},\mathbf y^{(i)}\right)\right) = \frac{1}{m}\sum_{k = 1}^m \mathds{1}_{\left(y_k^{(i)} \neq \hat{y}_k^{(i)}\right)}
$$
If one label is predicted incorrectly, this accounts for an error of $\frac{1}{m}$.
For a test set $D_{\text{test}}$ the hamming loss becomes:
$$
\text{HammingLoss}(C, D_{\text{test}}) = \frac{1}{n} \sum_{i = 1}^n \frac{1}{m}\sum_{k = 1}^m \mathds{1}_{\left(y_k^{(i)} \neq \hat{y}_k^{(i)}\right)}
$$

The following measures are scores instead of loss function like the two previous ones.

\item[(iii)] The accuracy, also called Jaccard-Index, for a test set $D_{\text{test}}$ is defined as:
$$
\text{accuracy}(C,D_{\text{test}}) = \frac{1}{n} \sum_{i = 1}^n \frac{\sum_{k = 1}^m \mathds{1}_{\left(y_k^{(i)} = 1 \text{ and } \hat{y}_k^{(i)}=1\right)}}{\sum_{k = 1}^m \mathds{1}_{\left(y_k^{(i)} = 1 \text{ or }\hat{y}_k^{(i)} = 1\right)}}
$$

\item[(iv)] The precision for a test set $D_{\text{test}}$ is defined as:

$$
\text{precision}(C,D_{\text{test}}) = \frac{1}{n} \sum_{i = 1}^n \frac{\sum_{k = 1}^m \mathds{1}_{\left(y_k^{(i)} = 1 \text{ and } \hat{y}_k^{(i)}=1\right)}}{\sum_{k = 1}^m \mathds{1}_{\left(\hat y_k^{(i)} = 1\right)}}
$$

\item[(v)] The recall for a test set $D_{\text{test}}$ is defined as:
$$
\text{recall}(C,D_{\text{test}}) = \frac{1}{n} \sum_{i = 1}^n \frac{\sum_{k = 1}^m \mathds{1}_{\left(y_k^{(i)} = 1 \text{ and } \hat{y}_k^{(i)}=1\right)}}{\sum_{k = 1}^m \mathds{1}_{\left(y_k^{(i)} = 1\right)}}
$$

\item[(vi)] For a test set $D_{\text{test}}$ the F$_1$-index is defined as follows:
$$
\text{F}_1(C,D_{\text{test}}) = \frac{1}{n} \sum_{i = 1}^n  \frac{2\sum_{k = 1}^m \mathds{1}_{\left(y_k^{(i)} = 1 \text{ and } \hat{y}_k^{(i)}=1\right)}}{\sum_{k = 1}^m \left(\mathds{1}_{\left(y_k^{(i)} = 1\right)} +  \mathds{1}_{\left(\hat y_k^{(i)} = 1\right)}\right)}
$$
The F$_1$-index is the harmonic mean of recall and precision on a per instance basis. 

\end{itemize}

All these measures lie between 0 and 1. In the case of the subset $0/1$ loss and the hamming loss the values should be low, in all other cases the scores should be high. 
Demonstrative definitions with sets instead of vectors can be seen in \citet{Charte2015}.

\section{Implementation}

In this section, we briefly describe how to perform multilabel classifications in \pkg{mlr}. We provide small code examples for better illustration. 
A short tutorial is also available at \url{http://mlr-org.github.io/mlr-tutorial/release/html/multilabel/index.html}. 
The first step is to transform the multilabel dataset into a \samp{data.frame} in R. The columns must consist of vectors of features and 
one logical vector for each label that indicates if the label is present for the observation or not. 
To fit a multilabel classification algorithm in \pkg{mlr}, a multilabel task has to be created, where a vector of targets corresponding to the column names of the labels has to be specified. This task is an 
S3 object that contains the data, the target labels and further descriptive information. 
In the following example, the yeast data frame is extracted from the yeast.task, which is provided by the \pkg{mlr} package. 
Then the 14 label names of the targets are extracted and the multilabel task is created. 

\begin{example}
yeast = getTaskData(yeast.task)
labels = colnames(yeast)[1:14]
yeast.task = makeMultilabelTask(id = "multi", data = yeast, target = labels)
\end{example}

\subsection{Problem transformation methods}

To generate a problem transformation method learner, a binary classification base learner has to be created with \samp{makeLearner}. 
A list of available learners for classifications in \pkg{mlr} can be seen at \url{http://mlr-org.github.io/mlr-tutorial/release/html/integrated_learners/}.
Specific hyperparameter settings of the base learner can be set in this step through the \samp{par.vals} argument in \samp{makeLearner}. 
Afterwards, a learner for any problem transformation method can be created by applying the function \samp{makeMultilabel[$\hdots$]Wrapper}, where \texttt{[$\hdots$]} has to be
substituted by the desired problem transformation method. 
In the following example, two multilabel variants with rpart as base learner are created. The base learner is configured to output probabilities instead 
of discrete labels during prediction. 

\begin{example}
lrn = makeLearner("classif.rpart", predict.type = "prob")
multilabel.lrn1 = makeMultilabelBinaryRelevanceWrapper(lrn)
multilabel.lrn2 = makeMultilabelNestedStackingWrapper(lrn)
\end{example}

\subsection{Algorithm adaptation methods}

Algorithm adaptation method learners can be created directly with \samp{makeLearner}. The names of the specific learner can be looked up at  
\url{http://mlr-org.github.io/mlr-tutorial/release/html/integrated_learners/} in the multilabel section. 

\begin{example}
multilabel.lrn3 = makeLearner("multilabel.rFerns")
multilabel.lrn4 = makeLearner("multilabel.randomForestSRC")
\end{example}

\subsection{Train, predict and evaluate}
Training and predicting on data can be done as usual in \pkg{mlr} with the functions \samp{train} and \samp{predict}. 
Learner and task have to be specified in \samp{train}; trained model and task or new data have to be specified in \samp{predict}. 
\begin{example}
mod = train(multilabel.lrn1, yeast.task, subset = 1:1500)
pred = predict(mod, task = yeast.task, subset = 1501:1600)
\end{example}

The performance of the prediction can be assessed via the function \samp{performance}. Measures are represented as S3 objects and multiple objects can
be passed in as a list. 
The default measure for multilabel classification is the hamming loss (\textit{multilabel.hamloss}). 
All available measures for multilabel classification can be shown by \samp{listMeasures} or looked up in the appendix of the tutorial page\footnote{In the \pkg{mlr} package \textit{precision} is named \textit{positive predictive value} and  \textit{recall} is named \textit{true positive rate}.}
(\url{http://mlr-org.github.io/mlr-tutorial/release/html/measures/index.html}).

\begin{example}
performance(pred, measures = list(multilabel.hamloss, timepredict))
multilabel.hamloss        timepredict 
0.230              0.174 
listMeasures("multilabel")
# [1] "multilabel.ppv" "timepredict"         "multilabel.hamloss" "multilabel.f1"      
# [5] "featperc"       "multilabel.subset01" "timeboth"           "timetrain"          
# [9] "multilabel.tpr" "multilabel.acc"    
\end{example}

\subsection{Resampling} 

To properly evaluate the model, a resampling strategy, for example k-fold cross-validation, should be applied. 
This can be done in \pkg{mlr} by using the function \samp{resample}. 
First, a description of the subsequent resampling strategy, in this case three-fold cross-validation, is defined with \samp{makeResampleDesc}. 
The resample is executed by a call to the \samp{resample} function. The hamming loss is calculated for the binary relevance method. 

\begin{example}
rdesc = makeResampleDesc(method = "CV", stratify = FALSE, iters = 3)
r = resample(learner = multilabel.lrn1, task = yeast.task, resampling = rdesc, 
measures = list(multilabel.hamloss), show.info = FALSE)
r
# Resample Result
# Task: multi
# Learner: multilabel.classif.rpart
# multilabel.hamloss.aggr: 0.23
# multilabel.hamloss.mean: 0.23
# multilabel.hamloss.sd: 0.00
# Runtime: 6.36688
\end{example}

\subsection{Binary performance}

To calculate a binary performance measure like, e.g., the accuracy, the mean misclassification error (mmce) or the AUC for each individual label, 
the function \samp{getMultilabelBinaryPerformances} can be used. This function can be applied to a single multilabel test set prediction and also on 
a resampled multilabel prediction. To calculate the AUC, predicted probabilities are needed. These can be obtained by setting the argument \samp{predict.type = "prob"} in the \samp{makeLearner} function.

\begin{example}
head(getMultilabelBinaryPerformances(r$pred, measures = list(acc, mmce, auc)))
#        acc.test.mean mmce.test.mean auc.test.mean
# label1     0.7389326      0.2610674     0.6801810
# label2     0.5908151      0.4091849     0.5935160
# label3     0.6512205      0.3487795     0.6631469
# label4     0.6921804      0.3078196     0.6965552
# label5     0.7517584      0.2482416     0.6748458
# label6     0.7343815      0.2656185     0.6054968
\end{example}

\subsection{Parallelization}

In the case of a high number of labels and larger datasets, parallelization in the training and prediction process of the multilabel methods 
can reduce computation time.
This can be achieved by using the 
package \texttt{parallelMap} in \texttt{mlr} (see also the tutorial section of parallelization: \url{http://mlr-org.github.io/mlr-tutorial/release/html/multilabel/index.html}). 
Currently, only the binary relevance method is parallelizable, the classifier for each label is trained in parallel, as they are independent of each other. 
The other problem transformation methods will also be parallelizable (as far as possible) soon. 

\begin{example}
library(parallelMap)
parallelStartSocket(2)
lrn = makeMultilabelBinaryRelevanceWrapper("classif.rpart")
mod = train(lrn, yeast.task)
pred = predict(mod, yeast.task)
\end{example}

\section{Benchmark experiment}

\label{sec:bench}
In a similar fashion to \citet{Wang2014}, we performed a benchmark experiment on several datasets in order to compare the performances of the different multilabel algorithms. 

\textbf{Datasets}: 
In Table \ref{tab:bench} we provide an overview of the used datasets.  
We retrieved most datasets from the Mulan Java library for multilabel learning\footnote{\url{http://mulan.sourceforge.net/datasets-mlc.html}} as well as from other benchmark experiments of multilabel classification methods. See Table \ref{tab:bench} for article references. 
We uploaded all datasets to the open data platform OpenML \citep{Casalicchio2017, OpenML2013}, so they now can be downloaded directly from there.
In some of the used datasets, sparse labels had to be removed in order to avoid problems during cross-validation. 
Several binary classification methods have difficulties when labels are sparse, i.e., a strongly imbalanced binary target class can lead to constant predictions for that target. 
That can sometimes lead to direct problems in the base learners (when training on constant class labels is simply not allowed) or, e.g., in classifier chains, when the base learner cannot handle constant features.
Furthermore, one can reasonably argue that not much is to be learned for such a label.
Hence, labels that appeared in less than 2\% of the observations were removed. 
We computed \textit{cardinality} scores (based on the remaining labels) indicating the mean number of labels assigned to each case in the respective dataset. 
The following description of the datasets refers to the final versions after removal of sparse labels. 

\begin{itemize}

\item The first dataset (\textit{birds}) consists of 645 audio recordings of 15 different vocalizing bird species \citep{briggs2013new}. 
Each sound can be assigned to various bird species.

\item Another audio dataset (\textit{emotions}) consists of 593 musical files with 6 clustered emotional labels \citep{trohidis2008multi} and 72 predictors. 
Each song can be labeled with one or more of the labels $\lbrace$\textit{amazed-surprised}, \textit{happy-pleased}, \textit{relaxing-calm}, \textit{quiet-still}, \textit{sad-lonely}, 
\textit{angry-fearful}$\rbrace$. 

\item The \textit{genbase} dataset contains protein sequences that can be assigned to several classes of protein families \citep{diplaris2005protein}. 
The entire dataset contains 1186 binary predictors.

\item The \textit{langLog}\footnote{\url{http://languagelog.ldc.upenn.edu/nll/}} dataset includes 998 textual predictors and was originally compiled in the doctorial 
thesis of \citet{read2010scalable}. 
It consists of 1460 text samples that can be assigned to one or more topics such as \textit{language, politics, errors, humor} and \textit{computational linguistics}. 

\item The UC Berkeley \textit{enron}\footnote{\url{http://bailando.sims.berkeley.edu/enron_email.html}} dataset represents a subset of the original 
\textit{enron}\footnote{\url{http://www.cs.cmu.edu/~enron/}} dataset and consists of 1702 cases of emails with 24 labels and 1001 predictor 
variables \citep{Klimt2004}. 

\item A subset of the \textit{reuters}\footnote{\url{http://lamda.nju.edu.cn/data_MIMLtext.ashx}} dataset includes 2000 observations for text classification \citep{Zhang2008}.

\item The \textit{image}\footnote{\url{http://lamda.nju.edu.cn/data_MIMLimage.ashx}} benchmark dataset consists of $2000$ natural scene images. \citet{Zhou2007} extracted 135 features 
for each image and made it publicly available as \textit{processed} image dataset. Each observation can be associated with different label sets, where all possible labels are 
$\lbrace$\textit{desert}, \textit{mountains}, \textit{sea}, \textit{sunset}, \textit{trees}$\rbrace$. About $22\%$ of the images belong to more than one class. However, 
images belonging to three classes or more are very rare. 

\item The \textit{scene} dataset is an image classification task where labels like \textit{Beach, Mountain, Field, Urban} are assigned to each image \citep{Boutell20041757}.

\item The \textit{yeast} dataset \citep{NIPS2001_1964} consists of micro-array expression data, as well as phylogenetic profiles of yeast, and includes 2417 genes and 103 predictors. 
In total, 14 different labels can be assigned to a gene, but only 13 labels were used due to label sparsity. 

\item Another dataset for text-classification is the \textit{slashdot}\footnote{\url{http://slashdot.org}} dataset \citep{Read2011}. 
It consists of article titles and partial blurbs. Blurbs can be assigned to several categories (e.g., \textit{Science, News, Games}) based on word predictors. \\
\end{itemize}

\begin{table}[ht]
\centering
\adjustbox{width = 11cm}{
\centering
\begin{tabular}{lrrrrr}
\toprule
Dataset & Reference  & \# Inst.&\# Pred. & \# Labels & Cardinality \\ 
\midrule
birds* &\cite{briggs2013new}  &  645 & 260 & 15 & 0.96  \\ 
emotions & \cite{trohidis2008multi} & 593 & 72 & 6 & 1.87 \\ 
genbase* & \cite{diplaris2005protein} &  662 & 112 & 16 & 1.20  \\ 
langLog* & \cite{read2010scalable} & 1460 & 998 &  18 & 0.85 \\ 
enron* & \cite{Klimt2004} & 1702 & 1001 & 24 & 3.12 \\ 
reuters  & \cite{Zhang2008}  &2000 & 243 & 7 & 1.15 \\ 
image & \cite{Zhou2007} & 2000 & 135 & 5 & 1.24 \\ 
scene &\cite{Boutell20041757} & 2407 & 294 & 6 & 1.07 \\ 
yeast* & \cite{NIPS2001_1964} & 2417 & 103 &  13 & 4.22 \\ 
slashdot* & \cite{Read2011} & 3782 & 1079 & 14 & 1.13 \\ 
\bottomrule
\end{tabular}
}
\caption{Used benchmark datasets including number of instances, number of predictor, number of label and label cardinality. 
Datasets with an asterisk differ from the original dataset as sparse labels have been removed. The genbase dataset contained many constant factor variables, which were automatically removed by mlr.}
\label{tab:bench}
\end{table}

\textbf{Algorithms}: We used all multilabel classification methods currently implemented in \pkg{mlr}: 
binary relevance (BR), classifier chains (CC), nested stacking (NST), 
dependent binary relevance (DBR) and stacking (STA) as well as algorithm adaption methods of the 
\pkg{rFerns} (RFERN) and \pkg{randomForestSRC} (RFSRC) packages.
For DBR and STA the first level and meta level classifiers were equal. For CC and NST we chose random chain orders for each resample iteration. 

\textbf{Base Learners}: We employed two different binary classification base learner for each problem transformation algorithm: 
random forest (rf) of the \CRANpkg{randomForest} package \citep{Liaw2002} 
with $\text{ntree = 100}$ and adaboost (ad) from the \CRANpkg{ada} package \citep{Culp2012}, each with standard hyperparameter settings. 

\textbf{Performance Measures:} We used the six previously proposed performance measures. 
Furthermore, we calculated the reported values by means of a 10-fold cross-validation. 


\textbf{Code:} For reproducibility, the complete code and results can be downloaded from \citet{Figshare2016}. 
The R package \CRANpkg{batchtools} \citep{Bischl2015} was used for parallelization. 

\begin{figure}[!htb]
\begin{center}
  \includegraphics[width=\textwidth]{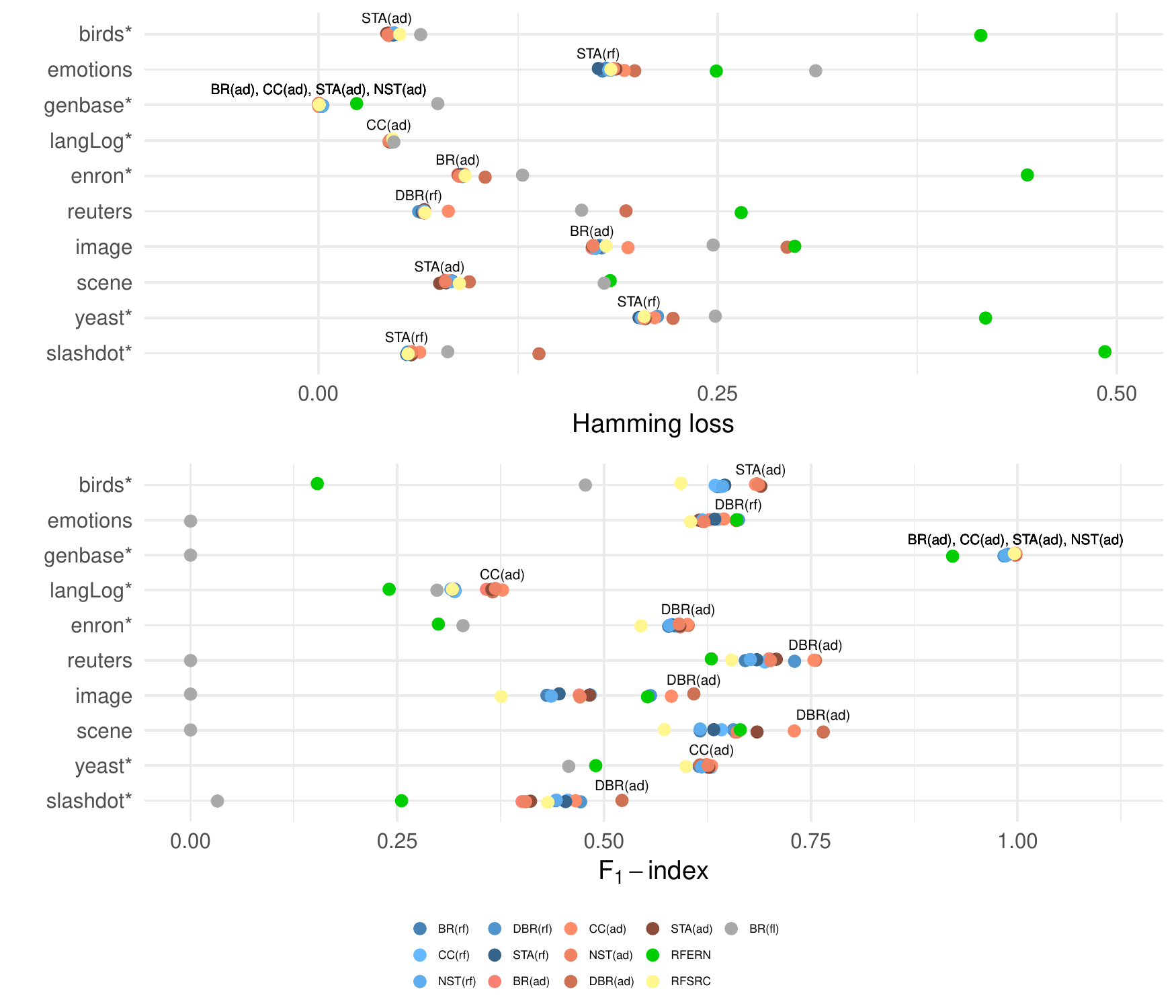}
  \caption[Results for hamming loss and F$_1$-index]{Results for hamming loss and F$_1$-index. The best performing algorithms are highlighted on the plot.}
  \label{fig:plotf1}
\end{center}
\end{figure}

The results for hamming loss and F$_1$-index are illustrated in Figure \ref{fig:plotf1}. Tables \ref{tab:benchmark2} and \ref{tab:benchmark3} contain performance values with the best performing algorithms highlighted in blue. For all remaining measures one may refer to the Appendix.
We did not perform any threshold tuning that would potentially improve some of the performance of the methods. 

The results of the problem transformation methods in this benchmark experiment concur with the general conclusions and results in \citet{Montanes2014}. 
The authors ran a similar benchmark study with penalized logistic regression as base learner. 
They concluded that, on average, DBR performs well in F$_1$ and accuracy. Also, CC outperform the other methods regarding the subset 0/1 loss most of the time. 
For the hamming loss measure they got mixed results, with no clear winner concordant to our benchmark results.
As base learner, on average, adaboost performs better than random forest in our benchmark study.

Considering the measure F$_1$, the problem transformation methods DBR, CC, STA and NST outperform RFERN and RFSRC on most of the datasets 
and also almost always perform better than BR, which does not consider dependencies among the labels. 
RFSRC and RFERN only perform well on either precision or recall, but in order to be considered as good classifiers they should perform well on both. 
The generally poor performances of RFERN can be explained by the working mechanism of the algorithm which randomly chooses variables and split points at each split of a fern. Hence, it cannot deal with too many features that are useless for the prediction of the target labels. 

\section{Summary}

In this paper, we describe the implementation of multilabel classification algorithms in the R package \pkg{mlr}. The problem transformation 
methods binary relevance, classifier chains, nested stacking, dependent binary relevance and stacking are implemented and can be used with any base learner that is accessible in \pkg{mlr}. 
Moreover, there is access to the multilabel classification versions of \mbox{\pkg{randomForestSRC}} and \pkg{RFerns}. 
We compare all of these methods in a benchmark experiment with several datasets and different implemented multilabel performance measures. 
The dependent binary relevance method performs well regarding the measures F$_1$ and \textit{accuracy}. 
Classifier chains outperform the other methods in terms of the subset 0/1 loss most of the time.
Parallelization is available for the binary relevance method and will be available soon for the other problem transformation methods. 
Algorithm adaptation methods and problem transformation methods that are currently not available can be incorporated in the current \pkg{mlr} framework easily. 
In our benchmark experiment we had to remove labels which occured too sparsely, because some algorithms crashed due to one class problems, which appeared during cross-validation. A solution to this problem and an implementation into the \pkg{mlr} framework is of great interest.

\FloatBarrier

\begin{table}[ht]
\centering
\scalebox{0.6}{
\begin{tabular}{rlllllllllllll}
  \hline
 & BR(rf) & CC(rf) & NST(rf) & DBR(rf) & STA(rf) & BR(ad) & CC(ad) & NST(ad) & DBR(ad) & STA(ad) & RFERN & RFSRC & BR(fl) \\ 
  \hline
\rowcolor[gray]{0.9}birds* & 0.0477 & 0.0479 & 0.0475 & 0.0472 & 0.0468 & 0.0442 & 0.0441 & 0.0436 & 0.0431 & \textcolor{blue}{\textbf{0.0429}} & 0.4148 & 0.0510 & 0.0641 \\ 
  emotions & 0.1779 & 0.1832 & 0.1818 & 0.1801 & \textcolor{blue}{\textbf{0.1753}} & 0.181 & 0.1916 & 0.1849 & 0.1981 & 0.1863 & 0.2492 & 0.1832 & 0.3114 \\ 
  \rowcolor[gray]{0.9}genbase* & 0.0021 & 0.0023 & 0.0025 & 0.0027 & 0.0023 & \textcolor{blue}{\textbf{0.0003}} & \textcolor{blue}{\textbf{0.0003}} & \textcolor{blue}{\textbf{0.0003}} & 0.0004 & \textcolor{blue}{\textbf{0.0003}} & 0.0240 & 0.0006 & 0.0748 \\ 
  langLog* & 0.0464 & 0.0465 & 0.0467 & 0.0464 & 0.0466 & 0.0451 & \textcolor{blue}{\textbf{0.0442}} & 0.0446 & 0.0447 & 0.0448 & 0.6673 & 0.0466 & 0.0473 \\ 
  \rowcolor[gray]{0.9}enron* & 0.0903 & 0.0904 & 0.0902 & 0.0909 & 0.0891 & \textcolor{blue}{\textbf{0.0874}} & 0.0913 & 0.0881 & 0.1045 & 0.0877 & 0.4440 & 0.0919 & 0.1279 \\ 
  reuters & 0.0663 & 0.0654 & 0.0661 & \textcolor{blue}{\textbf{0.0629}} & 0.065 & 0.0666 & 0.0814 & 0.0664 & 0.1926 & 0.0664 & 0.2648 & 0.0668 & 0.1649 \\ 
  \rowcolor[gray]{0.9}image & 0.1774 & 0.1791 & 0.1737 & 0.1761 & 0.1754 & \textcolor{blue}{\textbf{0.1714}} & 0.1939 & 0.1721 & 0.2935 & 0.1717 & 0.2983 & 0.1802 & 0.2472 \\ 
  scene & 0.0836 & 0.0809 & 0.0832 & 0.0796 & 0.0799 & 0.0791 & 0.0821 & 0.0796 & 0.0945 & \textcolor{blue}{\textbf{0.076}} & 0.1827 & 0.0884 & 0.1790 \\ 
  \rowcolor[gray]{0.9}yeast* & 0.2038 & 0.2044 & 0.2023 & 0.2123 & \textcolor{blue}{\textbf{0.2008}} & 0.2048 & 0.2105 & 0.2038 & 0.2221 & 0.2046 & 0.4178 & 0.2040 & 0.2486 \\ 
  slashdot* & 0.0558 & 0.0560 & 0.0559 & 0.0559 & \textcolor{blue}{\textbf{0.0554}} & 0.059 & 0.0635 & 0.0586 & 0.1382 & 0.0582 & 0.4925 & 0.0562 & 0.0811 \\ 
   \hline
\end{tabular}
}
\caption{Hamming loss} 
\label{tab:benchmark2}

\vspace{10mm}

\centering
\scalebox{0.6}{
\begin{tabular}{rlllllllllllll}
  \hline
 & BR(rf) & CC(rf) & NST(rf) & DBR(rf) & STA(rf) & BR(ad) & CC(ad) & NST(ad) & DBR(ad) & STA(ad) & RFERN & RFSRC & BR(fl) \\ 
  \hline
\rowcolor[gray]{0.9}birds* & 0.6369 & 0.6342 & 0.6433 & 0.64 & 0.6459 & 0.6835 & 0.683 & 0.6867 & 0.6846 & \textcolor{blue}{\textbf{0.6895}} & 0.1533 & 0.5929 & 0.4774 \\ 
  emotions & 0.6199 & 0.6380 & 0.6192 & \textcolor{blue}{\textbf{0.6625}} & 0.6337 & 0.6274 & 0.6449 & 0.6206 & 0.6598 & 0.615 & 0.6603 & 0.6046 & 0.0000 \\ 
  \rowcolor[gray]{0.9}genbase* & 0.9885 & 0.9861 & 0.9855 & 0.9835 & 0.9861 & \textcolor{blue}{\textbf{0.9977}} & \textcolor{blue}{\textbf{0.9977}} & \textcolor{blue}{\textbf{0.9977}} & 0.9962 & \textcolor{blue}{\textbf{0.9977}} & 0.9214 & 0.9962 & 0.0000 \\ 
  langLog* & 0.3192 & 0.3194 & 0.3148 & 0.3199 & 0.3167 & 0.3578 & \textcolor{blue}{\textbf{0.3772}} & 0.3686 & 0.3653 & 0.3643 & 0.2401 & 0.3167 & 0.2979 \\ 
  \rowcolor[gray]{0.9}enron* & 0.5781 & 0.5822 & 0.5791 & 0.5866 & 0.5826 & 0.592 & 0.6009 & 0.5906 & \textcolor{blue}{\textbf{0.6017}} & 0.5917 & 0.2996 & 0.5446 & 0.3293 \\ 
  reuters & 0.6708 & 0.6944 & 0.6769 & 0.7303 & 0.6846 & 0.6997 & 0.7537 & 0.7012 & \textcolor{blue}{\textbf{0.7556}} & 0.7082 & 0.6296 & 0.6541 & 0.0000 \\ 
  \rowcolor[gray]{0.9}image & 0.4308 & 0.4835 & 0.4362 & 0.5561 & 0.4456 & 0.47 & 0.5814 & 0.4709 & \textcolor{blue}{\textbf{0.6085}} & 0.4824 & 0.5525 & 0.3757 & 0.0000 \\ 
  scene & 0.6161 & 0.6420 & 0.6161 & 0.6563 & 0.6326 & 0.6585 & 0.73 & 0.661 & \textcolor{blue}{\textbf{0.765}} & 0.685 & 0.6647 & 0.5729 & 0.0000 \\ 
  \rowcolor[gray]{0.9}yeast* & 0.6148 & 0.6294 & 0.6180 & 0.6195 & 0.6244 & 0.6238 & \textcolor{blue}{\textbf{0.63}} & 0.6257 & 0.616 & 0.6266 & 0.4900 & 0.5991 & 0.4572 \\ 
  slashdot* & 0.4415 & 0.4562 & 0.4422 & 0.4716 & 0.4535 & 0.4009 & 0.4654 & 0.4052 & \textcolor{blue}{\textbf{0.5216}} & 0.411 & 0.2551 & 0.4320 & 0.0325 \\ 
   \hline
\end{tabular}
}
\caption{F$_1$-index} 
\label{tab:benchmark3}
\end{table}

\clearpage

\address{Philipp Probst\\
Department of Medical Informatics, Biometry and Epidemiology\\
LMU Munich\\
81377 Munich\\
Germany\\}
\email{probst@ibe.med.uni-muenchen.de}

\address{Quay Au\\
Department of Statistics\\
LMU Munich\\
80539 Munich\\
Germany\\}
\email{quay.au@stat.uni-muenchen.de}

\address{Giuseppe Casalicchio\\
Department of Statistics\\
LMU Munich\\
80539 Munich\\
Germany\\}
\email{giuseppe.casalicchio@stat.uni-muenchen.de}

\address{Clemens Stachl\\
Department of Psychology\\
LMU Munich\\
80802 Munich\\
Germany\\}
\email{clemens.stachl@psy.lmu.de}

\address{Bernd Bischl\\
Department of Statistics\\
LMU Munich\\
80539 Munich\\
Germany\\}
\email{bernd.bischl@stat.uni-muenchen.de}

\end{article}

\newpage
\appendix
\section{Appendices}

\begin{figure}[!htb]
\begin{center}
   \includegraphics[width=\textwidth]{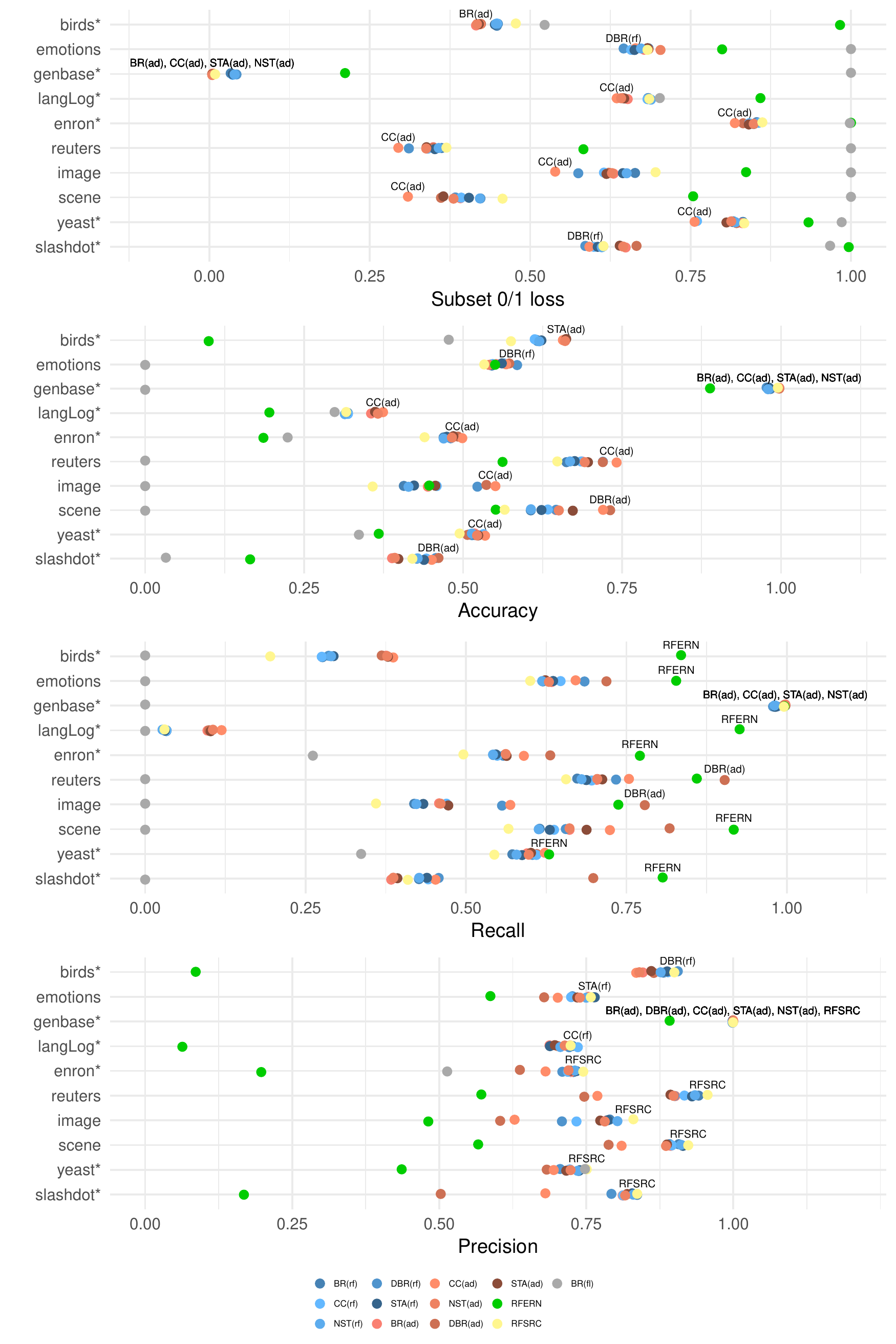} 
  \caption[Results for the remaining measures]{Results for the remaining measures.}
  \label{fig:plotprec}
\end{center}
\end{figure}

\clearpage

\begin{table}[ht]
\centering
\scalebox{0.6}{
\begin{tabular}{rlllllllllllll}
  \hline
 & BR(rf) & CC(rf) & NST(rf) & DBR(rf) & STA(rf) & BR(ad) & CC(ad) & NST(ad) & DBR(ad) & STA(ad) & RFERN & RFSRC & BR(fl) \\ 
  \hline
\rowcolor[gray]{0.9}birds* & 0.4481 & 0.4481 & 0.4466 & 0.4497 & 0.4451 & \textcolor{blue}{\textbf{0.4156}} & 0.4218 & 0.4171 & 0.4233 & 0.4202 & 0.9830 & 0.4777 & 0.5226 \\ 
  emotions & 0.6846 & 0.6575 & 0.6728 & \textcolor{blue}{\textbf{0.6457}} & 0.6626 & 0.6777 & 0.6643 & 0.7031 & 0.6845 & 0.6828 & 0.7992 & 0.6829 & 1.0000 \\ 
  \rowcolor[gray]{0.9}genbase* & 0.0333 & 0.0363 & 0.0393 & 0.0423 & 0.0363 & \textcolor{blue}{\textbf{0.0045}} & \textcolor{blue}{\textbf{0.0045}} & \textcolor{blue}{\textbf{0.0045}} & 0.0060 & \textcolor{blue}{\textbf{0.0045}} & 0.2115 & 0.0091 & 1.0000 \\ 
  langLog* & 0.6836 & 0.6829 & 0.6884 & 0.6842 & 0.6856 & 0.6521 & \textcolor{blue}{\textbf{0.6349}} & 0.6418 & 0.6438 & 0.6466 & 0.8589 & 0.6856 & 0.7021 \\ 
  \rowcolor[gray]{0.9}enron* & 0.8531 & 0.8413 & 0.8560 & 0.8408 & 0.8484 & 0.8496 & \textcolor{blue}{\textbf{0.819}} & 0.8484 & 0.8320 & 0.8408 & 1.0000 & 0.8619 & 0.9982 \\ 
  reuters & 0.3620 & 0.3405 & 0.3575 & 0.311 & 0.3515 & 0.349 & \textcolor{blue}{\textbf{0.2945}} & 0.338 & 0.3495 & 0.3385 & 0.5830 & 0.3695 & 1.0000 \\ 
  \rowcolor[gray]{0.9}image & 0.6635 & 0.6150 & 0.6505 & 0.575 & 0.6445 & 0.63 & \textcolor{blue}{\textbf{0.539}} & 0.6275 & 0.6225 & 0.619 & 0.8365 & 0.6955 & 1.0000 \\ 
  scene & 0.4225 & 0.3926 & 0.4217 & 0.3835 & 0.4046 & 0.3913 & \textcolor{blue}{\textbf{0.3095}} & 0.3805 & 0.3610 & 0.3648 & 0.7540 & 0.4570 & 1.0000 \\ 
  \rowcolor[gray]{0.9}yeast* & 0.8316 & 0.7600 & 0.8201 & 0.8167 & 0.8155 & 0.8304 & \textcolor{blue}{\textbf{0.7563}} & 0.8134 & 0.8217 & 0.806 & 0.9338 & 0.8337 & 0.9855 \\ 
  slashdot* & 0.6140 & 0.5994 & 0.6116 & \textcolor{blue}{\textbf{0.5859}} & 0.6052 & 0.6489 & 0.5923 & 0.6449 & 0.6658 & 0.6396 & 0.9966 & 0.6142 & 0.9675 \\ 
   \hline
\end{tabular}
}
\caption{Subset 0/1 loss} 
\label{tab:benchmark1}

\vspace{10mm}

\centering
\scalebox{0.6}{
\begin{tabular}{rlllllllllllll}
  \hline
 & BR(rf) & CC(rf) & NST(rf) & DBR(rf) & STA(rf) & BR(ad) & CC(ad) & NST(ad) & DBR(ad) & STA(ad) & RFERN & RFSRC & BR(fl) \\ 
  \hline
\rowcolor[gray]{0.9}birds* & 0.6153 & 0.6126 & 0.6197 & 0.6169 & 0.6232 & 0.6589 & 0.657 & 0.6604 & 0.6581 & \textcolor{blue}{\textbf{0.6621}} & 0.0999 & 0.5753 & 0.4774 \\ 
  emotions & 0.5453 & 0.5649 & 0.5464 & \textcolor{blue}{\textbf{0.5849}} & 0.5609 & 0.5519 & 0.5676 & 0.5408 & 0.5727 & 0.5427 & 0.5503 & 0.5332 & 0.0000 \\ 
  \rowcolor[gray]{0.9}genbase* & 0.9834 & 0.9806 & 0.9796 & 0.9773 & 0.9806 & \textcolor{blue}{\textbf{0.9972}} & \textcolor{blue}{\textbf{0.9972}} & \textcolor{blue}{\textbf{0.9972}} & 0.9957 & \textcolor{blue}{\textbf{0.9972}} & 0.8884 & 0.9950 & 0.0000 \\ 
  langLog* & 0.3185 & 0.3188 & 0.3140 & 0.3188 & 0.3161 & 0.3553 & \textcolor{blue}{\textbf{0.3741}} & 0.366 & 0.363 & 0.3615 & 0.1953 & 0.3161 & 0.2979 \\ 
  \rowcolor[gray]{0.9}enron* & 0.4693 & 0.4757 & 0.4694 & 0.4804 & 0.4742 & 0.483 & \textcolor{blue}{\textbf{0.4987}} & 0.4824 & 0.4919 & 0.4847 & 0.1859 & 0.4394 & 0.2241 \\ 
  reuters & 0.6625 & 0.6856 & 0.6682 & 0.7199 & 0.6754 & 0.6873 & \textcolor{blue}{\textbf{0.7414}} & 0.6912 & 0.7197 & 0.6964 & 0.5620 & 0.6482 & 0.0000 \\ 
  \rowcolor[gray]{0.9}image & 0.4068 & 0.4585 & 0.4142 & 0.5225 & 0.4228 & 0.4446 & \textcolor{blue}{\textbf{0.5508}} & 0.4458 & 0.5366 & 0.4564 & 0.4467 & 0.3578 & 0.0000 \\ 
  scene & 0.6064 & 0.6333 & 0.6067 & 0.6463 & 0.6233 & 0.646 & 0.7201 & 0.6505 & \textcolor{blue}{\textbf{0.7313}} & 0.6725 & 0.5513 & 0.5654 & 0.0000 \\ 
  \rowcolor[gray]{0.9}yeast* & 0.5091 & 0.5320 & 0.5138 & 0.514 & 0.5205 & 0.5182 & \textcolor{blue}{\textbf{0.5345}} & 0.522 & 0.5068 & 0.5239 & 0.3674 & 0.4945 & 0.3361 \\ 
  slashdot* & 0.4274 & 0.4421 & 0.4285 & 0.4569 & 0.4385 & 0.3883 & 0.4507 & 0.3925 & \textcolor{blue}{\textbf{0.4613}} & 0.3982 & 0.1651 & 0.4202 & 0.0325 \\ 
   \hline
\end{tabular}
}
\caption{Accuracy} 
\label{tab:benchmark4}

\vspace{10mm}

\centering
\scalebox{0.6}{
\begin{tabular}{rlllllllllllll}
  \hline
 & BR(rf) & CC(rf) & NST(rf) & DBR(rf) & STA(rf) & BR(ad) & CC(ad) & NST(ad) & DBR(ad) & STA(ad) & RFERN & RFSRC & BR(fl) \\ 
  \hline
\rowcolor[gray]{0.9}birds* & 0.2763 & 0.2752 & 0.2897 & 0.2859 & 0.2936 & 0.3755 & 0.3865 & 0.3772 & 0.3687 & 0.3784 & \textcolor{blue}{\textbf{0.8352}} & 0.1949 & 0.0000 \\ 
  emotions & 0.6197 & 0.6474 & 0.6187 & 0.6847 & 0.6358 & 0.6335 & 0.6708 & 0.6293 & 0.7189 & 0.6237 & \textcolor{blue}{\textbf{0.8276}} & 0.6001 & 0.0000 \\ 
  \rowcolor[gray]{0.9}genbase* & 0.9846 & 0.9819 & 0.9809 & 0.9786 & 0.9819 & \textcolor{blue}{\textbf{0.9977}} & \textcolor{blue}{\textbf{0.9977}} & \textcolor{blue}{\textbf{0.9977}} & 0.9962 & \textcolor{blue}{\textbf{0.9977}} & 0.9962 & 0.9955 & 0.0000 \\ 
  langLog* & 0.0334 & 0.0330 & 0.0270 & 0.0331 & 0.0308 & 0.0971 & 0.1191 & 0.1056 & 0.0995 & 0.102 & \textcolor{blue}{\textbf{0.9264}} & 0.0301 & 0.0000 \\ 
  \rowcolor[gray]{0.9}enron* & 0.5426 & 0.5487 & 0.5421 & 0.5580 & 0.5466 & 0.5611 & 0.5902 & 0.5619 & 0.6314 & 0.5633 & \textcolor{blue}{\textbf{0.771}} & 0.4959 & 0.2613 \\ 
  reuters & 0.6733 & 0.6959 & 0.6801 & 0.7338 & 0.6875 & 0.7038 & 0.754 & 0.7046 & \textcolor{blue}{\textbf{0.9032}} & 0.7123 & 0.8598 & 0.6559 & 0.0000 \\ 
  \rowcolor[gray]{0.9}image & 0.4192 & 0.4696 & 0.4228 & 0.5562 & 0.4335 & 0.4581 & 0.5691 & 0.4603 & \textcolor{blue}{\textbf{0.7787}} & 0.4724 & 0.7374 & 0.3598 & 0.0000 \\ 
  scene & 0.6148 & 0.6373 & 0.6134 & 0.6555 & 0.6306 & 0.6614 & 0.7243 & 0.6613 & 0.8174 & 0.6879 & \textcolor{blue}{\textbf{0.9173}} & 0.5662 & 0.0000 \\ 
  \rowcolor[gray]{0.9}yeast* & 0.5722 & 0.6097 & 0.5788 & 0.6035 & 0.5874 & 0.5951 & 0.6229 & 0.5978 & 0.6104 & 0.6013 & \textcolor{blue}{\textbf{0.6296}} & 0.5442 & 0.3365 \\ 
  slashdot* & 0.4267 & 0.4412 & 0.4270 & 0.4574 & 0.4391 & 0.3834 & 0.4526 & 0.3868 & 0.6984 & 0.3931 & \textcolor{blue}{\textbf{0.8065}} & 0.4094 & 0.0000 \\ 
   \hline
\end{tabular}
}
\caption{Recall} 
\label{tab:benchmark5}

\vspace{10mm}

\centering
\scalebox{0.6}{
\begin{tabular}{rlllllllllllll}
  \hline
 & BR(rf) & CC(rf) & NST(rf) & DBR(rf) & STA(rf) & BR(ad) & CC(ad) & NST(ad) & DBR(ad) & STA(ad) & RFERN & RFSRC & BR(fl) \\ 
  \hline
\rowcolor[gray]{0.9}birds* & 0.8812 & 0.8889 & 0.8764 & \textcolor{blue}{\textbf{0.9056}} & 0.8874 & 0.8461 & 0.8349 & 0.8401 & 0.8648 & 0.8605 & 0.0859 & 0.8996 &  \\ 
  emotions & 0.7627 & 0.7242 & 0.7499 & 0.7265 & \textcolor{blue}{\textbf{0.7644}} & 0.7537 & 0.7014 & 0.739 & 0.6783 & 0.7347 & 0.5869 & 0.7577 &  \\ 
  \rowcolor[gray]{0.9}genbase* & 0.9987 & 0.9987 & 0.9987 & 0.9987 & 0.9987 & \textcolor{blue}{\textbf{0.9995}} & \textcolor{blue}{\textbf{0.9995}} & \textcolor{blue}{\textbf{0.9995}} & \textcolor{blue}{\textbf{0.9995}} & \textcolor{blue}{\textbf{0.9995}} & 0.8917 & \textcolor{blue}{\textbf{0.9995}} &  \\ 
  langLog* & 0.7267 & \textcolor{blue}{\textbf{0.7356}} & 0.7058 & 0.7207 & 0.6882 & 0.6874 & 0.7228 & 0.7133 & 0.7014 & 0.6965 & 0.0632 & 0.7233 &  \\ 
  \rowcolor[gray]{0.9}enron* & 0.7283 & 0.7188 & 0.7305 & 0.7092 & 0.7331 & 0.7235 & 0.6807 & 0.7198 & 0.6371 & 0.7233 & 0.1973 & \textcolor{blue}{\textbf{0.7448}} & 0.5135 \\ 
  reuters & 0.9411 & 0.9168 & 0.9346 & 0.8995 & 0.9298 & 0.9014 & 0.7689 & 0.8983 & 0.7465 & 0.8931 & 0.5715 & \textcolor{blue}{\textbf{0.9562}} &  \\ 
  \rowcolor[gray]{0.9}image & 0.7899 & 0.7333 & 0.8029 & 0.7086 & 0.7865 & 0.7841 & 0.6281 & 0.7814 & 0.6036 & 0.7737 & 0.4813 & \textcolor{blue}{\textbf{0.83}} &  \\ 
  scene & 0.9071 & 0.8956 & 0.9112 & 0.8917 & 0.9143 & 0.8936 & 0.81 & 0.8856 & 0.7879 & 0.8872 & 0.5662 & \textcolor{blue}{\textbf{0.9233}} &  \\ 
  \rowcolor[gray]{0.9}yeast* & 0.7372 & 0.7218 & 0.7351 & 0.7055 & 0.7389 & 0.7225 & 0.6947 & 0.7233 & 0.6827 & 0.7159 & 0.4361 & \textcolor{blue}{\textbf{0.7508}} & 0.7478 \\ 
  slashdot* & 0.8365 & 0.8127 & 0.8298 & 0.7927 & 0.8277 & 0.8119 & 0.6804 & 0.8161 & 0.5025 & 0.8196 & 0.1679 & \textcolor{blue}{\textbf{0.8366}} &  \\ 
   \hline
\end{tabular}
}
\caption[prec]{Precision \footnotemark} 
\label{tab:benchmark6}
\end{table}

\footnotetext{For the featureless learner we have no precision results for several datasets. The reason is that the featureless learner does not predict any value in all observations in these datasets. Hence, the denominator in the precision formula is always zero. \textit{mlr} predicts NA in this case.}

\end{article}

\end{document}